\documentclass{ieeeaccess}
\usepackage{amsmath,amssymb,amsfonts}
\usepackage{algorithmic}
\usepackage{graphicx}
\usepackage{textcomp}
\usepackage{url}
\usepackage[table]{xcolor}
\usepackage{biblatex}
\usepackage{textpos}

\usepackage{booktabs}
\usepackage{tabularx}
\usepackage{colortbl}
\usepackage{textcomp}
\usepackage{verbatim}
\usepackage{multirow}
\usepackage{makecell}
\usepackage{subcaption}
\usepackage{siunitx}
\usepackage[super]{nth}
\usepackage{inputenc}
\usepackage{xspace}
\usepackage[T1]{fontenc}
\usepackage{pifont}

\usepackage[breaklinks=true,
            letterpaper=true,
            colorlinks,
            citecolor = cyan,
            urlcolor = blue,
            bookmarks=false]{hyperref}
\usepackage[capitalize]{cleveref}
\usepackage{float}

\usepackage{pgfplots}
\pgfplotsset{compat=1.17}
\NewSpotColorSpace{PANTONE}
  \AddSpotColor{PANTONE} {PANTONE3015C} {PANTONE\SpotSpace 3015\SpotSpace C} {1 0.3 0 0.2}
  \SetPageColorSpace{PANTONE}

\definecolor{LightGray}{gray}{0.95}
\definecolor{applegreen}{rgb}{0.55, 0.9, 0.0}
\definecolor{amber}{rgb}{1.0, 0.75, 0.0}
\definecolor{darkpastelgreen}{rgb}{0.01, 0.75, 0.24}

\newcommand{\textapprox}{\raisebox{0.5ex}{\texttildelow}}
\newcommand{\ie}{\emph{i.e.}}
\newcommand{\eg}{\emph{e.g.}}

\expandafter\let\csname c@tblerows\endcsname\rownum

\def\BibTeX{{\rm B\kern-.05em{\sc i\kern-.025em b}\kern-.08em
    T\kern-.1667em\lower.7ex\hbox{E}\kern-.125emX}}

\begin{document}

\history{Date of publication May 05, 2022.}
\doi{10.1109/ACCESS.2022.3172939}

\title{Image Classification with Small Datasets: Overview and Benchmark}

\author{\uppercase{Lorenzo Brigato}\authorrefmark{1},
\uppercase{Bj{\"o}rn Barz}\authorrefmark{2}, 
\uppercase{Luca Iocchi}\authorrefmark{1},
\uppercase{and Joachim Denzler}\authorrefmark{2}
}
\address[1]{Department
of Computer, Control and Management Engineering, Sapienza University of Rome, Italy (e-mail: brigato@diag.uniroma.it)}
\address[2]{Department of Mathematics and Computer Science, Friedrich Schiller University Jena, Germany (e-mail: bjoern.barz@uni-jena.de)}

\markboth
{L. Brigato \headeretal: Image Classification with Small Datasets: Overview and Benchmark}
{L. Brigato \headeretal: Image Classification with Small Datasets: Overview and Benchmark}

\corresp{Corresponding author: Lorenzo Brigato (e-mail: brigato@diag.uniroma1.it)}


\begin{abstract}
Image classification with small datasets has been an active research area in the recent past.
However, as research in this scope is still in its infancy, two key ingredients are missing for ensuring reliable and truthful progress: a systematic and extensive overview of the state of the art, and a common benchmark to allow for objective comparisons between published methods.
This article addresses both issues.
First, we systematically organize and connect past studies to consolidate a community that is currently fragmented and scattered.
Second, we propose a common benchmark that allows for an objective comparison of approaches.
It consists of five datasets spanning various domains (\eg, natural images, medical imagery, satellite data) and data types (RGB, grayscale, multispectral).
We use this benchmark to re-evaluate the standard cross-entropy baseline and ten existing methods published between 2017 and 2021 at renowned venues.
Surprisingly, we find that thorough hyper-parameter tuning on held-out validation data results in a highly competitive baseline and highlights a stunted growth of performance over the years.
Indeed, only a single specialized method dating back to 2019 clearly wins our benchmark and outperforms the baseline classifier.
\end{abstract}


\titlepgskip=-15pt

\maketitle

\section{Introduction}
\label{sec:intro}

\PARstart{M}{any}  recent advances in computer vision and machine learning in general have been achieved by large-scale pre-training on massive datasets \cite{dosovitskiy2021vit,cui2018large,radford2019gpt}.
For instance, the most popular dataset for image classification, ImageNet-1k \cite{russakovsky2015imagenet}, contains one thousand classes, each comprised of several hundred or over one thousand training examples.
However, reaching high recognition performance by training on large-scale datasets is strictly connected to the laborious process of collecting and labeling large quantities of samples.
Application scenarios in which the usual pre-training on large web-sourced image datasets is useless due to a strong domain shift (\eg, document style classification \cite{stutzmann2016clamm}) or even impossible due to different data modalities (\eg, multi-channel spectral data from satellites \cite{helber2019eurosat}) strictly depend on methods for learning directly from the limited amounts of data available.

\emph{Deep learning from small datasets} is a research area that has been receiving increasing interest in the past couple of years \cite{oyallon2017scaling, barz2020deep, brigato2021close, bruintjes2021vipriors}.
The distinctive feature that makes this field of research differ from other learning problems is that, the use of additional, external datasets, \eg, for pre-training neural networks, is not available.
While nowadays popular datasets contain hundreds or thousands of training examples per class, \emph{deep learning from small data} trains neural networks on tens or hundreds of samples per category.
These extreme settings exacerbate the well-known weaknesses of neural networks \ie, being prone to memorizing spurious correlations among training features instead of actually learning a general function for the requested task \cite{geirhos2020shortcut}.
Therefore, deploying a performant classifier in this scenario remains an important challenge. 

To keep focus, we limit the scope of our study to image classification with few examples, excluding other computer vision domains such as object detection or semantic segmentation that have also recently gained increasing attention \cite{liu2019generative, bruintjes2021vipriors}.
Despite their invaluable importance, such domains are still lacking a sufficiently large body of literature for an extensive overview.
In contrast, our literature analysis on image classification with small datasets led to a substantial number of existing works, probably because image classification remains one of the most established tasks for artificial neural networks. 

A key missing piece of the current literature is an objective comparison of proposed methods due to the lack of a common benchmark.
Fortunately, there recently have been activities to establish common benchmarks and organize challenges to foster direct competition between proposed methods \cite{bruintjes2021vipriors}.
Still, they are often limited to a single dataset, \eg, ImageNet \cite{russakovsky2015imagenet}, which comprises a different type of data than usually encountered in a small-data scenario.
Moreover, most existing works compare their proposed method against insufficiently tuned baselines \cite{barz2020deep} or baselines trained with default hyper-parameters \cite{oyallon2017scaling,ulicny2019harmonic,kayhan2020translation,sun2020visual,kobayashi2021t}, which makes it easy to outperform them.
Careful hyper-parameter optimization (HPO) \cite{bischl2021hpo} is not only crucial for applying deep learning techniques in practice but also for a fair comparison between different methods so that each can obtain its optimal or near-to-optimal performance.
Comparing against an untuned baseline with default hyper-parameters does not provide clear evidence
of improvements.
Additionally, due to the fragmentation of the relevant literature, approaches are rarely compared to the existing state of the art, obfuscating the progress of the field.

The contributions of this paper enrich the literature on image classification with small datasets with two fundamental building blocks that are currently missing: 1) a review of the recent literature and 2) a dedicated benchmark.
The former represents the first comprehensive collection of works on image classification with small datasets.
We provide a clear overview of the current literature and existing approaches.
The second building block is a dedicated benchmark allowing for a direct, objective, and informative comparison of existing and future methods.
The benchmark consists of five datasets from different domains: natural images of everyday objects, fine-grained classification, medical imagery, satellite images, and handwritten documents.
Two datasets consist of non-RGB data, where the common large-scale pre-training and the fine-tuning procedure is not straightforward, emphasizing the need for methods that can learn from limited amounts of data from scratch.
Our dataset splits, implementations of all compared methods, and code for reproducing our experiments is publicly available under \url{https://github.com/lorenzobrigato/gem}.

\begin{figure}[t]
    \resizebox{\linewidth}{!}{
            \input{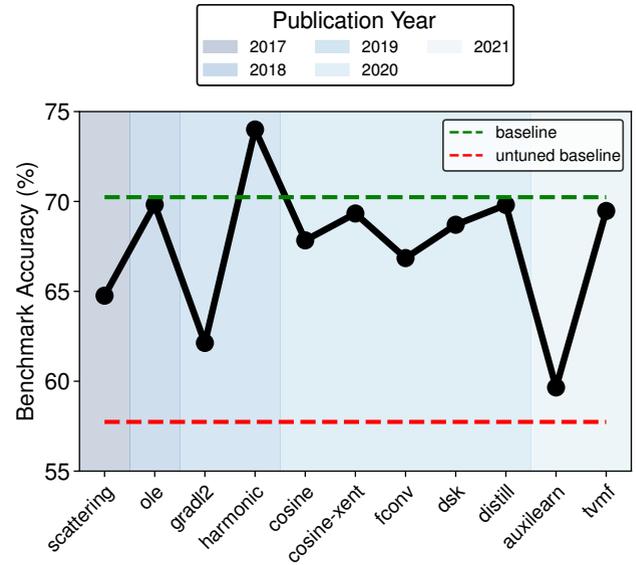}}
    \caption{Accuracy of state-of-the-art methods and baselines on the proposed benchmark. The untuned baseline (red dashed line) is trained with default hyper-parameters \ie, a learning rate of 0.1, and weight decay of \(10^{-4}\). Conversely, for all other methods, including the baseline (green dashed line), we performed hyper-parameter optimization.
    Methods are ordered on the x-axis according to their publication year.}
    \label{fig:benchyear}
\end{figure}

For our benchmark, we carefully optimize the hyper-parameters of all methods for each dataset individually on held-out validation data, while evaluating the final performance on a separate test split.
Surprisingly and somewhat disillusioning, we discover two key findings that are summarized in \cref{fig:benchyear}: 1) hyper-parameter optimization makes the categorical cross-entropy loss a strong baseline that is outperformed by only one of the ten specialized methods evaluated; 2) there is no clear performance progress considering the approaches published in the recent literature.
The untuned baseline (red dashed line), \ie, a classifier trained with default hyper-parameters, underperforms the baseline trained with HPO (green dashed line) by \textapprox13 percentage points.
In more detail, such a strong baseline is obtainable by combining small batch sizes (\ie, as small as 8), strong regularization (\ie, weight decays up to \textapprox\(10^{-2}\)), and small learning rates confined to a limited range (\ie, \textapprox\([10^{-4}, 10^{-3}]\)).
To show the lack of performance progress, we order the evaluated methods on the proposed benchmark along the x-axis of \cref{fig:benchyear} according to the original publication year.
We notice that the accuracy does not monotonically increase throughout the years, as it would be desirable.
   
Hence, we learn that default hyper-parameters found in the image classification literature are inadequate in data-deficient scenarios and should be substituted with properly tuned configurations, \eg, more aggressive regularization for baseline classifiers.
We identify the lack of such hyper-parameter tuning and comprehensive comparisons in the existing literature as the cause for the illusion of frequent progress.
In contrast, only a single method analyzed in our study is able to outperform the baseline consistently in a realistic setting.

First, we differentiate the faced learning settings from related research areas in \cref{sec:reldomains}.
We then dive deeper into image classification from small datasets and review the existing literature in this field in \cref{sec:survey}.
Afterward, we describe our proposed benchmark, starting with the methods selected for the comparison in \cref{sec:methods} and datasets in \cref{sec:datasets}.
Our experimental setup and training procedure are detailed in \cref{sec:setup} and the results are presented in \cref{sec:results}.
In \cref{sec:discussion}, we discuss potential limitations of our comparison.
\Cref{sec:conclusions} summarizes the conclusions from our study.

\section{Related Research Areas}
\label{sec:reldomains}

To avoid potential sources of confusion, we will first give a brief description of research areas that are related to \emph{learning from small datasets} but different in crucial aspects.

\emph{Transfer learning} is a well-established approach that uses knowledge from a previous task to solve a secondary, generally related task \cite{pan2009survey, weiss2016survey}.
Often, the target task has a much smaller number of training examples, hence, fine-tuning a pre-trained network has benefits in terms of recognition performance.
Note that this area should not be included in the literature on image classification with small datasets, as the pre-training is performed on large image databases such as ImageNet.
This is not possible in some scenarios, especially when the type of the input data (\eg, multispectral images from satellites) is different from RGB.

\emph{ Domain adaptation } is a subfield of \emph{transfer learning} that assumes related source and target tasks, with the latter undergoing a distributional shift.
Typically, the target task has only unlabeled data or a few annotated pairs.
We refer the reader to \cite{wang2018deep} for a survey on this topic.
Similar to the previous paradigm, \emph{domain adaptation} also uses knowledge extraction from a data-rich source task.

\emph{ Few-shot learning } is a domain that has received considerable attention over the past years \cite{wang2020generalizing,hospedales2021meta}.
The goal of this approach is to train a model that learns to recognize similarities and in turn perform tasks in data-poor target domains, including scenarios with only 1 or 5 samples per class.
While the goal of \emph{few-shot learning} overlaps with that of \emph{learning from small datasets}, their implementations practically differ.
\emph{Few-shot learning} relies on a qualitatively rich base set of annotated pairs, from which it can \emph{meta-learn} more general representations that are then used to solve the few-shot task.
On the other hand, \emph{learning from small datasets} uses a very modest training set to learn from that is slightly larger than a few shots, but does not have access to any large-scale pre-training data.

\emph{ Weakly supervised learning } deals with training datasets with few, noisy or inconsistent labels \cite{zhou2018brief}.
Moreover, in this domain, there are no assumptions about the dimension of the training dataset.
There could be samples for which there is no training label, but which are still available to a semi-supervised algorithm.
In contrast, \emph{learning from small datasets} assumes that the correct label is available for each member of the small training dataset.

\emph{Long-tailed recognition}, also known as \emph{unbalanced classification}, is a largely researched area, as high-class imbalance naturally occurs in many classification problems \cite{johnson2019survey}.
In this scenario, the learner is regularized to not only learn effective representations to classify the majority classes but also correctly recognize the minority classes \cite{cui2019class}.
The number of samples in the tail of the training distribution is comparable in size to the dimensions of the datasets used in \emph{classification with small data}.
However, the latter excludes the existence of other classes with a large number of samples, which would contribute to the learning of general representations.

\begin{figure}[t]
    \resizebox{\linewidth}{!}{
            \input{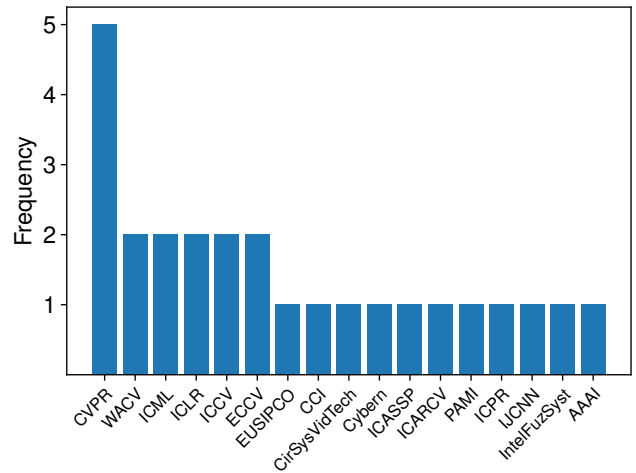}}
    \caption{Distribution of publication venues concerning the reviewed body of literature.}
    \label{fig:venues}
\end{figure}

\section{Literature Review}
\label{sec:survey}

In this section, we present the body of literature that we found after a careful search for image classification methods with small datasets.
Moreover, we propose a classification taxonomy to organize the relevant research directions that have been explored so far.

\subsection{Search} 

We included in our collection mainly articles that fulfill two criteria:
First, they should have been peer-reviewed at renowned conferences and journals.
Second, they propose an approach for specifically tackling the problem of learning from a small sample.
To verify the latter, we checked whether the experiments described in each candidate paper were executed on small or sub-sampled versions of popular image classification datasets.

To find relevant papers, we searched popular search engines and archives using keyword arguments that strongly match the features of this research domain such as \emph{data efficiency}, \emph{small data}, \emph{small datasets} and \emph{data-efficient}.
Along with that, we also used direct paper references as a channel to find additional connections.

As a result of our search, we found 26 articles published between 2015 and 2021.  
Five of these works are published in journals while the rest have been presented at conferences or workshops.
\Cref{fig:venues} shows the distribution of articles among venues.
From the figure, it stands out that computer vision conferences (\eg, CVPR, ICCV, ECCV, WACV) are the venues where the community is more inclined to publish.
Instead, machine learning (\eg, ICLR and ICML) and signal processing conferences are slightly behind in terms of preferences.

\begin{figure}[t]
    \includegraphics[width=\linewidth]{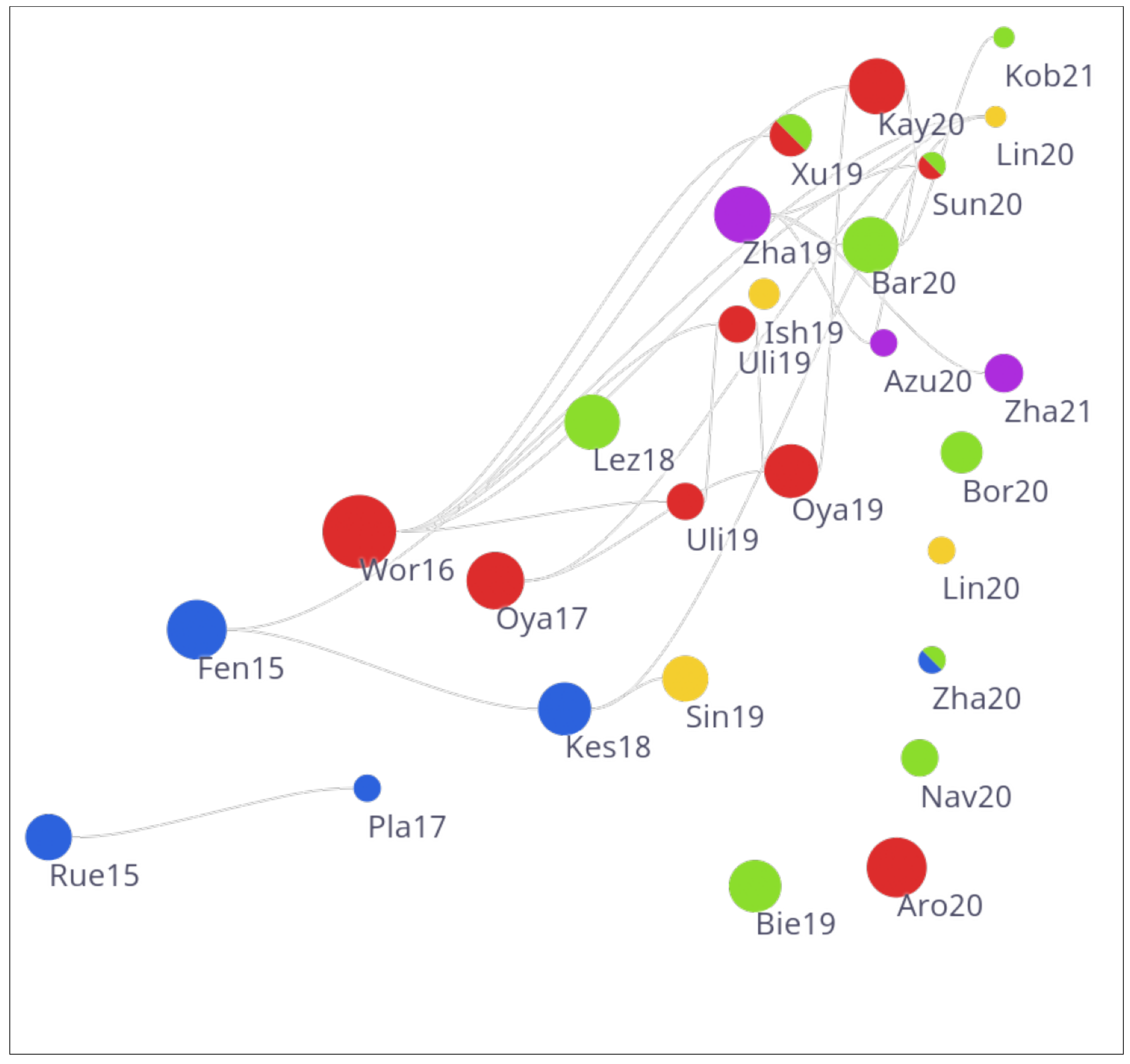}
    \caption{Visualization of relative connections within the reported body of literature. Works are shown chronologically along the x-axis and distributed along the y-axis for better readability. Colors represent the taxonomic classes \emph{architecture} (\textcolor{red}{$\bullet$}), \emph{cost function} (\textcolor{applegreen}{$\bullet$}), \emph{data augmentation} (\textcolor{violet}{$\bullet$}), \emph{latent augmentation} (\textcolor{amber}{$\bullet$}), and \emph{warm-starting} (\textcolor{blue}{$\bullet$}).}
    \label{fig:literature}
\end{figure}

\subsection{Map}

To provide a clear visual picture of the literature and extract additional insights on the current status of the research domain, we employ the online tool Litmaps\footnote{\url{https://www.litmaps.co/}} to generate a literature map of the state-of-the-art methods for image classification with small datasets (\cref{fig:literature}).

Each circle on this map represents a paper, denoted with a short keyword composed of part of the first-author name and the year of publication.
The radius of the circle is proportional to the total number of citations and the colors represent the taxonomic classes to which the paper has been assigned by our breakdown.
Therefore, circles of the same color represent articles that proposed methods belonging to the same methodological approach.
More details about the taxonomy will follow in the next sub-section.
Furthermore, articles are ordered along the x-axis according to publication time in ascending order.
Connections between circles indicate references.

From the literature map, two things seem evident: the domain has gained increasing interest in the recent past and exhibits low connectivity in terms of references.
The first fact, undoubtedly positive, is evincible from the growth of circles in the rightmost part of the map.
While only 7 papers appeared between 2015 and 2018, the number of publications has grown to 19 in the subsequent three years.
The second factor, which is instead negative, regards the fragmentation of the current state of the art.
We notice that different papers published in 2020 or 2021 do not reference any of the older papers (bottom right corner of \cref{fig:literature}) indicating that a comparison between newly proposed approaches and the existing state of the art is often missing.
Moreover, articles that are graphed to the upper part of the map, despite having higher connectivity, still did not reference multiple previous works.
For instance, the most cited work by other members of the map, denoted with \emph{Wor16} and corresponding to \cite{worrall2017harmonic}, has only been referenced by 5 subsequent papers. 
In response to the aforementioned connectivity issue of the current literature, we propose and organize the largest collection of works related to image classification with small datasets.

\subsection{Taxonomy}
\label{subsec:taxonomy}

In this section, we categorize current methods into a methodological taxonomy that well fits the existing literature.
Precisely, we distinguish five families of approaches, depending on how the model is regularized:

\begin{itemize}
    \item \emph{architecture} (\textcolor{red}{$\bullet$})
    \item \emph{cost function} (\textcolor{applegreen}{$\bullet$})
    \item \emph{data augmentation} (\textcolor{violet}{$\bullet$})
    \item \emph{latent augmentation} (\textcolor{amber}{$\bullet$})
    \item \emph{warm-starting} (\textcolor{blue}{$\bullet$})
\end{itemize}

We briefly illustrate the intrinsic characteristics of each taxonomic category and describe the papers belonging to each one.
Notice that our proposed taxonomy classifies methods along independent axes, but it is also possible that some papers proposed a combination of multiple components falling into different families.
A trivial example could be the proposal of a novel architecture trained with a new loss function.

\subsubsection{Architecture (\textcolor{red}{$\bullet$})}

This category includes all possible modifications to standard network architectures at all topological levels. 
This covers changes applied to network layers, blocks, stages, etc.
Reasonably, this is a quite general class that is composed of multiple sub-categories and contains a significant number of works in our collection (7).

An important contribution to this family derives from those methods relying on \emph{geometric priors} \ie, techniques that use intrinsic and geometric approaches coming from the signal-processing literature.
We include the use of pre-set, fixed filters based on wavelet transformations \cite{oyallon2017scaling, oyallon2018scattering} or discrete cosine transform \cite{ulicny2019harmonic, ulicny2019harmonicnet}. 
Also, invariance to rotation and translation are desirable properties that have been embedded into CNNs through the use of steerable filters or circular harmonics \cite{worrall2017harmonic}. 

Other architectural changes aim for invariance with respect to input transformations without employing mathematics from signal processing.
For instance, F-Conv \cite{kayhan2020translation} uses an alternative padding strategy to decrease image-boundary effects and improve translation invariance.
Xu et al.~\cite{xu2020towards} introduced an alternative convolution block to favor the learning of scale-invariant representations.
Similarly, Sun et al.~\cite{sun2020visual} modified the standard residual block and proposed a multi-branch structure with anti-aliasing modules and selective kernels.
Finally, Arora et al.~\cite{arora2019harnessing} employ neural tangent kernels, which are equivalent to infinitely wide networks.

\subsubsection{Cost function (\textcolor{applegreen}{$\bullet$})} 
\label{subsubsec:1}
Networks are optimized to minimize a so-called \emph{cost function} to learn the target task.
Cost functions are generally composed of multiple items, including different regularization or penalty terms (\eg, weight decay).
We prefer to employ the term \emph{cost} instead of \emph{loss} since, colloquially, the latter is widely used to exclusively indicate the error between predictions and targets without considering other terms.  
We notice that this family of regularization is the most popular within this community with 9 different contributions.

Two works proposed losses based on cosine similarity to regularize network training \cite{barz2020deep, kobayashi2021t}, while Sun et al.~\cite{sun2020visual} employed a three-term cost function including a change of the standard cross-entropy to a combination of itself and the cosine loss.
Xu et al.~\cite{xu2020towards} hand-crafted a rotation-invariant regularizer based on prior knowledge, which is added to the cost function as a penalty term.
Bietti et al.~\cite{bietti2019kernel} tested multiple existing regularization principles (\ie, gradient penalties, spectral norms) and also proposed new regularization penalties for learning from small datasets.
Moreover, Lezama et al.~\cite{lezama2018ole} conceived a geometric loss term based on an orthogonal low-rank embedding that can be plugged into any cost function and encourages embeddings for different classes to be orthogonal.
Bornschein et al.~\cite{bornschein2020small} calibrate the cross-entropy loss using a scalar temperature parameter, which is optimized alternatingly on a validation set. Navon et al.~\cite{navon2021auxiliary} use implicit differentiation to either learn how to combine multiple losses or predict auxiliary targets.
Also overlapping with this category is the work of Zhao et al.~\cite{zhao2020distilling}, who employ a two-stage training using a modified version of the contrastive loss for self-supervised representation learning and a distillation loss to regularize the features learned by the final classifier.

\subsubsection{Data augmentation (\textcolor{violet}{$\bullet$})}
All methods that increase the size of the training dataset reside in this category. 
For instance, standard data augmentation (\ie, cropping, flipping transformations, etc.) is a classical strategy belonging to this family.
We also include in this category the coupling of generative models with classifiers since, the latter, can be used to synthesize additional examples and improve classification accuracy.
Data augmentation techniques have received plenty of attention in the deep learning literature \cite{shorten2019survey} and are based on simple tricks \cite{zhong2020random} or more complex automated strategies \cite{cubuk2019autoaugment, cubuk2020randaugment}.

Surprisingly, in our literature collection, we only found 3 works proposing data augmentation approaches specifically designed for the small-sample regime.
One common approach consists in deep adversarial augmentation \cite{zhang2019dada, zhang2020deep} and the other one on generative latent implicit conditional optimization \cite{azuri2021generative}.

\subsubsection{Latent augmentation (\textcolor{amber}{$\bullet$})} 
Stochastic or adversarial transformations applied to features inside networks constitute the main building block of this family of regularizers.

Ishii and Sato \cite{ishii2019training} adversarially augment features at randomly selected hidden layers by adding small perturbations to the original features extracted from training data.
Lin et al.~\cite{lin2020efficient,lin2020latent} propose a framework that regularizes the classifier by sampling latent variables encoded in Gaussian distributions.
Finally, Keshari et al.~\cite{keshari2019guided} introduce a more advanced dropout policy by measuring the strength of each node.

\subsubsection{Warm-starting (\textcolor{blue}{$\bullet$})} 
This class of approaches employs algorithmic schemes to initialize the classifier with weights that favor better learning on small datasets.
We assume that warm-starting may happen a single time or multiple times in the training process.
For instance, the \emph{self-supervised} paradigm belongs to this family since encoders are first pre-trained on unsupervised tasks and then used to warm-start the final classifier. 
 
The oldest work of our collection \cite{rueda2015supervised}, and its successive implementation \cite{plata2017effective}, trained networks in a layer-wise greedy manner, analogous to that used in unsupervised deep networks.
In other words, each layer is initialized with the weights obtained from the precedent run.
Similarly, Feng and Darrell \cite{feng2015learning} proposed a multi-step initialization algorithm that adapts the model complexity to the available training data and learns the structure of filters.
Subsequent research decouples the structure and strength of convolutional filters to reduce the overall number of parameters by using a dictionary-based filter learning algorithm and, subsequently, standard training \cite{keshari2018learning}. 
More recently, Zhao et al.~\cite{zhao2020distilling} used self-supervised learning to learn a general encoder followed by self-distillation coupled with the standard classification objective.

\section{Benchmark Methods}
\label{sec:methods}

In this section, we first present in more detail the approaches that we evaluated on our benchmark (\cref{subsec:methods_eval}).
Then, we name the methods that we discarded and motivate such exclusions by describing their limiting factors (\cref{subsec:methods_disc}).

\subsection{Evaluated}
\label{subsec:methods_eval}

Along with a baseline cross-entropy classifier, we evaluated ten specialized methods of our literature review.
Out of the ten approaches, four belong to the taxonomic category \emph{architecture} (\textcolor{red}{$\bullet$}), five to \emph{cost function} (\textcolor{applegreen}{$\bullet$}), and one to both \emph{warm-starting} and \emph{cost function} (\textcolor{blue}{$\bullet$}\textcolor{applegreen}{$\bullet$}). 
We describe the evaluated approaches in more detail in the following.

\subsubsection{Cross-Entropy Loss} This is the widely used standard loss function for classification. We use it as a baseline with standard network architectures and optimization algorithms.

\subsubsection{Deep Hybrid Networks (DHN) (\textcolor{red}{$\bullet$})} 
This approach represents one of the first attempts to incorporate pre-defined geometric priors via a hybrid approach of combining pre-defined and learned representations \cite{oyallon2017scaling, oyallon2018scattering}.
According to the authors, decreasing the number of parameters to learn could make deep networks more data-efficient, especially in settings where the scarcity of data would not allow the learning of low-level feature extractors.
Deep hybrid networks first perform a scattering transform on the input image generating feature maps and then apply standard convolutional blocks.  
The spatial scale of the scattering transform is controlled by the parameter \(J \in \mathbb{N}\).

\subsubsection{Orthogonal Low-Rank Embedding (OLÉ) (\textcolor{applegreen}{$\bullet$})} 
Lezama et al. \cite{lezama2018ole} proposed this geometric loss that is intended to reduce intra-class variance and enforce inter-class margins for deep networks.
This method collapses deep features into a
learned linear subspace, or union of them, and inter-class subspaces are pushed to be as orthogonal as possible. The contribution of the low-rank embedding to the overall loss is weighted by the hyper-parameter \(\lambda_\mathrm{ole}\).

\subsubsection{Grad-\(\mathbf{\ell_{2}}\) Penalty (\textcolor{applegreen}{$\bullet$})} 
This is a regularization strategy tested in the context of improving generalization on small datasets \cite{bietti2019kernel}.
The \(\ell_{2}\) (squared) gradient norm is computed for the input samples and used as a penalty in the loss weighted by parameter \(\lambda_\mathrm{grad}\).
Among many regularization approaches evaluated in by Bietti et al.~\cite{bietti2019kernel}, we have chosen the grad-\(\ell_{2}\) penalty because it was among the best-performing methods in the experiments with ResNet and sub-sampled versions of CIFAR-10.

\subsubsection{Cosine Loss (\textcolor{applegreen}{$\bullet$})} Barz and Denzler \cite{barz2020deep} proposed this loss to decrease overfitting in problems with scarce data.
Thanks to an $\ell_2$ normalization of the learned feature space, the cosine loss is invariant against scaling of the network output and solely focuses on the directions of feature vectors instead of their magnitude.
In contrast to the softmax function used with the cross-entropy loss, the cosine loss does not push the activations of the true class towards infinity, which is commonly considered a cause of overfitting \cite{szegedy2016rethinking,he2018bag}.
A further increase in performance was obtained by combining the cosine with the cross-entropy loss after an additional layer on top of the embeddings learned with the cosine loss.

\subsubsection{Harmonic Networks (HN) (\textcolor{red}{$\bullet$})} 
HN uses a set of preset filters based on windowed cosine transform at several frequencies which are combined by learnable weights \cite{ulicny2019harmonicnet, ulicny2019harmonic}.
Similar to hybrid networks, the idea of the harmonic block is to have a useful geometric prior that can help to avoid overfitting.
Harmonic networks use Discrete Cosine Transform filters which have excellent energy compaction properties and are widely used for image compression.

\subsubsection{Full Convolution (F-Conv) (\textcolor{red}{$\bullet$})} Kayhan et al. \cite{kayhan2020translation} proposed F-Conv to improve the translation invariance of convolutional filters.
Standard CNNs exploit image boundary effects and learn filters that can exploit the absolute spatial locations of objects in images.
In contrast, full convolution applies each value in the filter to all values in the image.
According to Kayhan et al.~\cite{kayhan2020translation}, improving translation invariance strengthens the visual inductive prior of convolution, leading to increased data efficiency in the small-data setting.

\subsubsection{Dual Selective Kernel Networks (DSKN) (\textcolor{red}{$\bullet$})}
In this approach \cite{sun2020visual}, the standard residual block is modified, keeping the skip connection, with two forward branches that use \(\ 1 \times 1\) convolutions, selective kernels \cite{li2019selective} and an anti-aliasing module.
To further regularize training, only one of the two branches is randomly selected in the forward and backward passes, while at inference, the two paths are weighted equally.

Besides the specialized network architecture, the original work uses a combination of three custom loss functions \cite{sun2020visual}.
Despite best efforts, we were unable to derive the correct implementation from the ambiguous description of these loss functions in the paper.
Therefore, we only use the DSKN architecture with the standard cross-entropy loss.

\subsubsection{Distilling Visual Priors (DVP) (\textcolor{blue}{$\bullet$}\textcolor{applegreen}{$\bullet$})} 
Zhao et al. \cite{zhao2020distilling} introduce this two-stage framework that, firstly, learns a teacher model via self-supervised learning using the popular MoCo approach \cite{chen2020improved}, and secondly, distills the representations into a student classifier using self-distillation \cite{heo2019comprehensive}.
A contribution of this work regards the novel margin loss implemented to better learn general representations under the data-deficient scenario.
Hence, in addition to the hyper-parameters for MoCo, a margin $\lambda_m$ needs to be set.
In addition, for the second stage of training, $\lambda_{dist}$ weights the contribution of the distillation loss to the overall cost function.

\subsubsection{Auxiliary Learning (AuxiLearn) (\textcolor{applegreen}{$\bullet$})}  
AuxiLearn is a method for generating meaningful and novel auxiliary tasks \cite{navon2021auxiliary}.
This is achieved by training an auxiliary network to generate auxiliary labels while training another, primary network to learn both the original task and the auxiliary task.
The objective is to push the representation of the primary network to generalize better on the main task by exploiting multi-task learning as a regularizer.
To train this method, multiple hyper-parameters need to be set.
First, the dimension of the auxiliary set which is a small percentage of the training data.
Then, the strength of the auxiliary loss component along with the update-frequency of auxiliary gradients.
Finally, the type of the auxiliary network \ie, linear or not linear, and, for the latter case, its depth and number of units per layer.

\subsubsection{T-vMF Similarity (\textcolor{applegreen}{$\bullet$})} 
This similarity \cite{kobayashi2021t} is a generalization of the cosine similarity and was proposed to make modern CNNs more robust to some realistic learning situations such as class imbalance, few training samples, and noisy labels.
As the name suggests, T-vMF Similarity is mainly based on the von Mises-Fisher distribution of directional statistics and built on top of the heavy-tailed student-t distribution.

The combination of these two ingredients provides high compactness in high-similarity regions and low similarity in heavy-tailed ones.
The degree of compactness/dispersion of the similarity is controlled by the parameter \(\kappa\).

\subsection{Discarded}
\label{subsec:methods_disc}

We now describe the approaches belonging to the literature overview that we discarded and provide the related reasons.

A group of papers does not provide the original implementation of the proposed methods.
To avoid unreliable or wrong re-evaluations, we restricted our final choice to approaches for which source code was available.
For this reason, we discarded two contributions from \emph{architecture} (\textcolor{red}{$\bullet$}) \cite{arora2019harnessing,xu2020towards}, three from \emph{cost function} (\textcolor{applegreen}{$\bullet$}) \cite{bornschein2020small,xu2020towards,sun2020visual}, two from \emph{latent augmentation} (\textcolor{amber}{$\bullet$}) \cite{ishii2019training, keshari2019guided}, and four from \emph{warm-starting} (\textcolor{blue}{$\bullet$}) \cite{rueda2015supervised, feng2015learning, plata2017effective, keshari2018learning}.
Only for DSKN \cite{sun2020visual}, we were able to implement the proposed architecture by ourselves but unable to correctly derive the proposed loss function.

In a few cases, despite best efforts and the availability of source code, we were unable to reproduce a properly working implementation.
Faced obstacles included divergence issues \cite{lin2020efficient,lin2020latent} or very poor results  \cite{worrall2017harmonic, zhang2019dada, zhang2020deep}.
Finally, one \emph{data augmentation} method employs a pre-trained VGG network to compute a perceptual loss embedded in the proposed framework \cite{azuri2021generative}.
This approach does not fully respect the assumptions of deep learning from small data, which does not allow external data or pre-trained models.
Furthermore, it can not be adapted straightforwardly to different data types (e.g., grayscale or multi-spectral images).

\section{Benchmark Datasets}
\label{sec:datasets}

\begin{figure}[t]
    \resizebox{\linewidth}{!}{
            \input{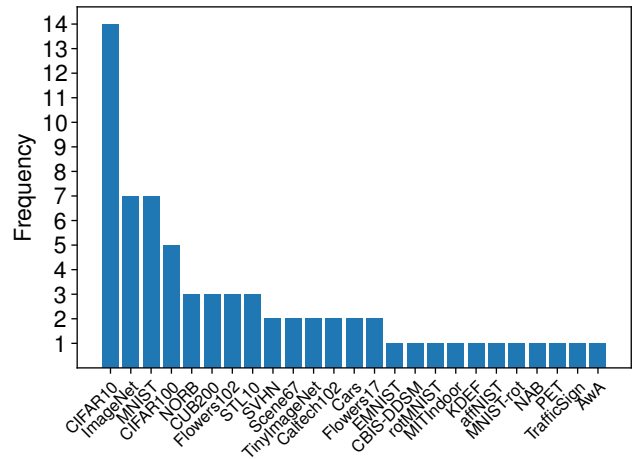}}
    \caption{Frequency of datasets used used in the reviewed literature.}
    \label{fig:datasets-freq}
\end{figure}

Most works on deep learning from small datasets use custom sub-sampled versions of popular standard image classification benchmarks such as ImageNet \cite{russakovsky2015imagenet} or CIFAR \cite{krizhevsky2009learning}.
This is visible from \cref{fig:datasets-freq}, where we show the frequency of use of all datasets employed in the reviewed body of literature.
CIFAR-10, ImageNet, and CIFAR-100 were utilized 14, 7, and 5 times, respectively, accounting for an overall fraction of \textapprox40\% of datasets together.
It is also striking that \textapprox50\% of the articles carried out experiments on CIFAR-10, making it the most frequently used dataset in this community.

This limited variety bears the risk of overfitting research progress to individual datasets and the domain covered by them, in this case, photographs of natural scenes and everyday objects.
In particular, this is not the domain typically dealt with in a small-data scenario, where specialized data that is difficult to obtain or annotate is in the focus.
Additionally, very recent work showed that high performance on ImageNet does not necessarily correlate with high performance on other vision datasets \cite{tuggener2021enough}.

Therefore, we compile a diverse benchmark consisting of five datasets from a variety of domains and with different data types and numbers of classes.
We sub-sampled all datasets to fit the small-data regime, except for CUB \cite{wah2011cub}, which was already small enough.
By default, we aimed for 50 training images per class.
To account for variance stemming from the sub-sampling operation, we employ 3 different sets of dataset splits.
In other words, we sub-sample the original datasets three times, when possible, and train and evaluate methods on each sub-portion independently.
Full training splits are only used for the final training and split into training (\textapprox 60\%) and validation sets (\textapprox 40\%) for hyper-parameter optimization.
For testing the final models trained on the \emph{trainval\{i\}} splits, with \(\emph{i} \in \{0,1,2\}\), we used official standard test datasets where they existed.
Only for two datasets, namely EuroSAT \cite{helber2019eurosat} and ISIC 2018 \cite{codella2019isic}, we had to create own test splits, \ie, \emph{test\{i\}}.
Given the already small size of CUB, we could not vary the training splits but only the \emph{train} and \emph{val} ones. 
A summary of the dataset statistics is given in \cref{tab:datasets} and \cref{fig:dataset-examples} shows examples from all datasets.
In the following, we briefly describe each dataset used for our benchmark.

\subsubsection{ciFAIR-10} 
Barz and Denzler proposed a variant \cite{barz2020cifair} of the popular CIFAR-10 dataset \cite{krizhevsky2009learning}, which comprises low-resolution images of size $32 \times 32$ from 10 different classes of everyday objects.
To a large part, its popularity stems from the fact that the low image resolution allows for fast training of neural networks and hence rapid experimentation.
However, the test dataset of CIFAR-10 contains about 3.3\% duplicates from the training set \cite{barz2020cifair}, which can potentially bias the evaluation.
The ciFAIR-10 dataset \cite{barz2020cifair} provides a variant of the test set, where these duplicates have been replaced with new images from the same domain.

\subsubsection{Caltech-UCSD Birds-200-2011 (CUB)}
CUB is a fine-grained dataset of 200 bird species \cite{wah2011cub}.
Annotating this kind of images typically requires a domain expert and is hence costly.
Therefore, the dataset is rather small and only comprises 30 training images per class.
Pre-training on related large-scale datasets is hence the de-facto standard for CUB \cite{cui2018large,lin2015bilinear,Simon19:implicit,zheng2017learning}, which makes it particularly interesting for research on sample-efficient methods closing the gap between training from scratch and pre-training.

\subsubsection{ISIC 2018} 
ISIC 2018 is a medical dataset consisting of dermoscopic skin lesion images, annotated with one of seven possible skin disease types \cite{codella2019isic}.
Since medical data usually requires costly expert annotations, this domain is important to be covered by a benchmark on data-efficient deep learning.
Due to the small number of classes, we increase the number of images per class to 80 for this dataset, so that the size of the training set is more similar to our other datasets.

\subsubsection{EuroSAT} 
This is a multispectral image dataset based on Sentinel-2 satellite images of size \(64 \times 64\) covering 13 spectral bands \cite{helber2019eurosat}.
Each image is annotated with one of ten land cover classes.
This dataset does not only exhibit a substantial domain shift compared to standard pre-training datasets such as ImageNet but also a different number of input channels.
This scenario renders the standard pre-training and fine-tuning procedure impossible.

Nevertheless, Helber et al.~\cite{helber2019eurosat} adhere to this procedure by fine-tuning a CNN pre-trained on RGB images using different combinations of three out of the 13 channels of EuroSAT.
Unsurprisingly, they find that the combination of the R, G, and B channels provides the best performance in this setting.
This limitation to three channels due to pre-training is a waste of data and potential.
In our experiments on a smaller subset of EuroSAT, we found that using all 13 channels increases the classification accuracy by 9.5\% compared to the three RGB channels when training from scratch.

\subsubsection{CLaMM} 
CLaMM is a dataset for \textbf{C}lassification of \textbf{La}tin \textbf{M}edieval \textbf{M}anuscripts \cite{stutzmann2016clamm}.
It was originally used in the ICFHR 2016 Competition for Script Classification, where the task was to classify grayscale images of Latin scripts from handwritten books dated 500 C.E. to 1600 C.E. into one of twelve script style classes such as \emph{Humanistic Cursive}, \emph{Praegothica} etc.
This domain is quite different from that of typical pre-training datasets such as ImageNet and one can barely expect any useful knowledge to be extracted from ImageNet about medieval documents.
In addition, the standard pre-training and fine-tuning procedure would require converting the grayscale images to RGB for passing them through the pre-trained network, which incurs a waste of parameters.

\begin{table*}[t]
    \centering
    \renewcommand{\arraystretch}{1.3}
    \setlength{\tabcolsep}{10pt}
    \rowcolors{2}{LightGray}{white}
    \resizebox{\linewidth}{!}{
    \begin{tabular}{l c c c c c c l l}
        \toprule
         Dataset &  Classes & Imgs/Class & \#Trainval &  \#Test &  Train Splits & Test Splits & Problem Domain & Data Type \\
         \midrule
         ciFAIR-10 \cite{krizhevsky2009learning,barz2020cifair} &  10 &         50 &       500 &  10,000 & \emph{trainval\{i\}} & \emph{test}  & Natural Images & RGB (32x32) \\
         CUB \cite{wah2011cub}                                  &     200 &         30 &     5,994 &   5,794 & \emph{trainval} & \emph{test}  & Fine-Grained   & RGB \\
         ISIC 2018 \cite{codella2019isic}                       &       7 &         80 &       560 &   1,944 & \emph{trainval\{i\}} & \emph{test\{i\}}  & Medical        & RGB \\
         EuroSAT \cite{helber2019eurosat}                       &      10 &         50 &       500 &  19,500 & \emph{trainval\{i\}} & \emph{test\{i\}}  & Remote Sensing & Multispectral \\
         CLaMM \cite{stutzmann2016clamm}                        &      12 &         50 &       600 &   2,000 & \emph{trainval\{i\}} & \emph{test}  & Handwriting    & Grayscale \\
         \bottomrule
    \end{tabular}
    }
    \caption{Datasets constituting our benchmark. To account for variance stemming from the sub-sampling operation, we employ three different training splits (except for CUB).
    On ISIC 2018 and EuroSAT, given the lack of a fixed testing set, we also vary such splits. 
    The value of \emph{i} refers to the identifier of the three splits, \ie, \(\emph{i} \in \{0,1,2\}\).}
    \label{tab:datasets}
\end{table*}

\begin{figure*}[t]
    \setlength{\tabcolsep}{0pt}
    \renewcommand{\arraystretch}{0}
    \hspace{2.5cm}
    \begin{subfigure}[b]{.32\linewidth}%
        \offinterlineskip%
        \resizebox{\linewidth}{!}{%
            \includegraphics[height=5cm]{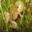}%
            \includegraphics[height=5cm]{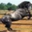}%
            \includegraphics[height=5cm]{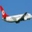}%
        }
        \resizebox{\linewidth}{!}{%
            \includegraphics[height=5cm]{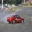}%
            \includegraphics[height=5cm]{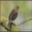}%
            \includegraphics[height=5cm]{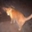}%
        }
        \caption{ciFAIR-10}
    \end{subfigure}%
    \hfill%
    \begin{subfigure}[b]{.288\linewidth}%
        \offinterlineskip%
        \resizebox{\linewidth}{!}{%
            \includegraphics[height=5cm]{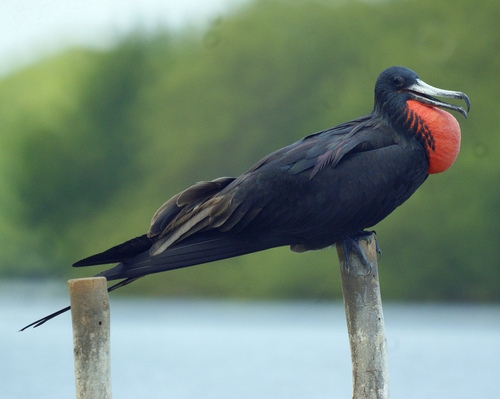}%
            \includegraphics[height=5cm]{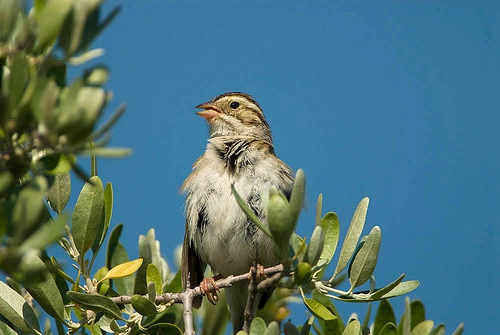}%
        }
        \resizebox{\linewidth}{!}{%
            \includegraphics[height=5cm]{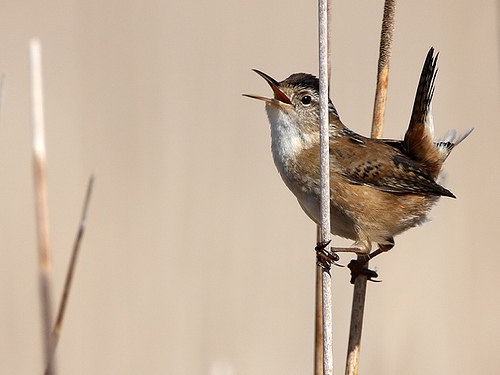}%
            \includegraphics[height=5cm]{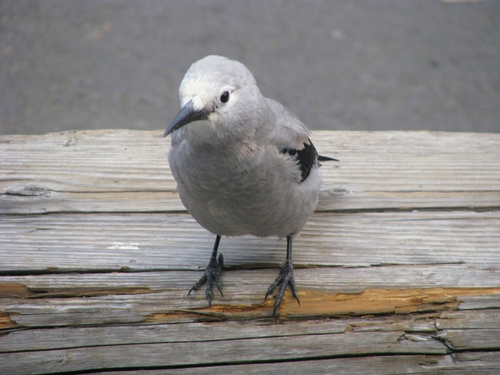}%
        }
        \caption{CUB}
    \end{subfigure}
    \hspace{2.5cm}
    \\[.8\baselineskip]
    \begin{subfigure}[b]{.3\linewidth}%
        \offinterlineskip%
        \resizebox{\linewidth}{!}{%
            \includegraphics[height=5cm]{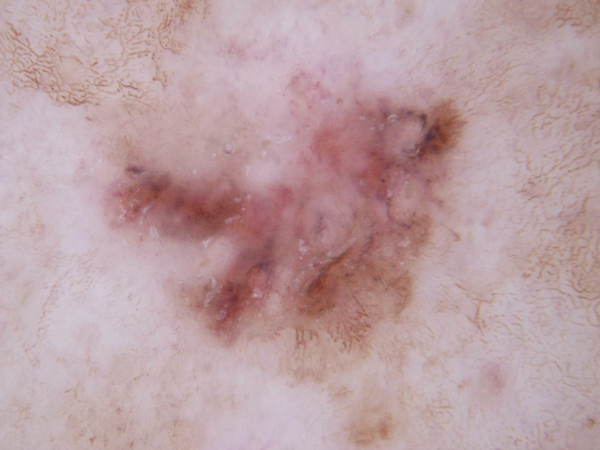}%
            \includegraphics[height=5cm]{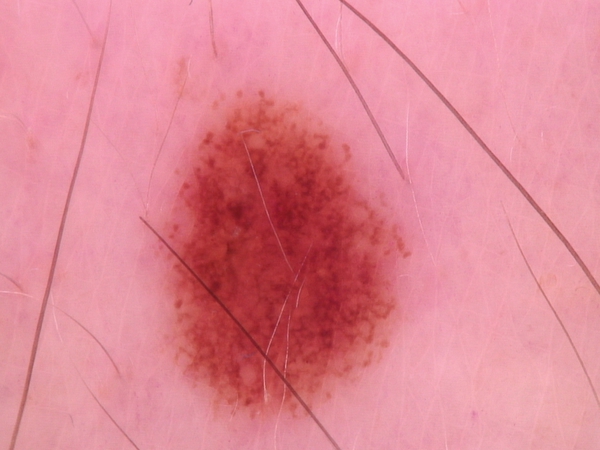}%
        }
        \resizebox{\linewidth}{!}{%
            \includegraphics[height=5cm]{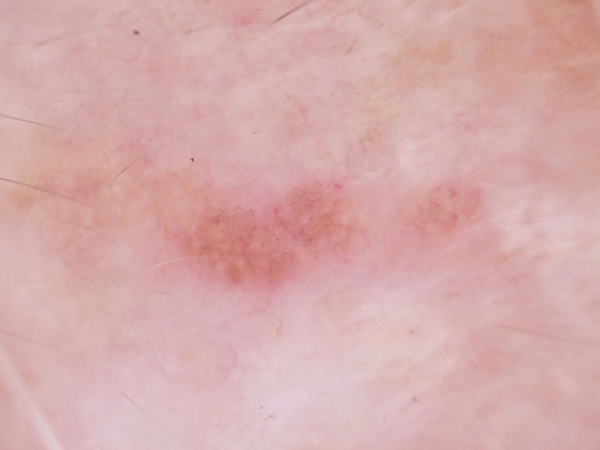}%
            \includegraphics[height=5cm]{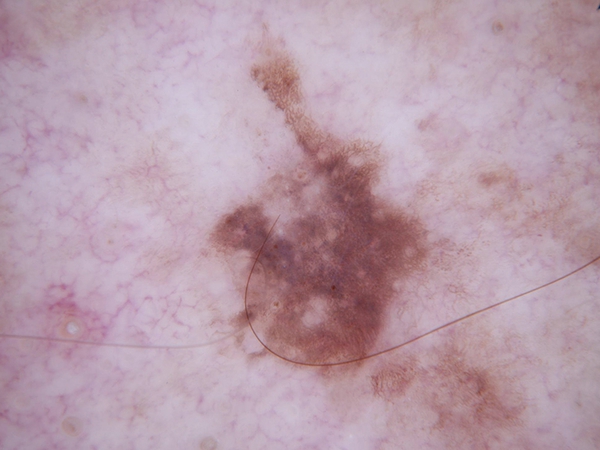}%
        }
        \caption{ISIC 2018}
    \end{subfigure}%
    \hfill%
    \begin{subfigure}[b]{.338\linewidth}%
        \offinterlineskip%
        \resizebox{\linewidth}{!}{%
            \includegraphics[height=5cm]{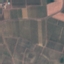}%
            \includegraphics[height=5cm]{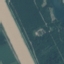}%
            \includegraphics[height=5cm]{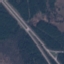}%
        }
        \resizebox{\linewidth}{!}{%
            \includegraphics[height=5cm]{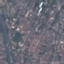}%
            \includegraphics[height=5cm]{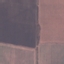}%
            \includegraphics[height=5cm]{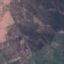}%
        }
        \caption{EuroSAT}
    \end{subfigure}%
    \hfill%
    \begin{subfigure}[b]{.309\linewidth}%
        \offinterlineskip%
        \resizebox{\linewidth}{!}{%
            \includegraphics[height=5cm]{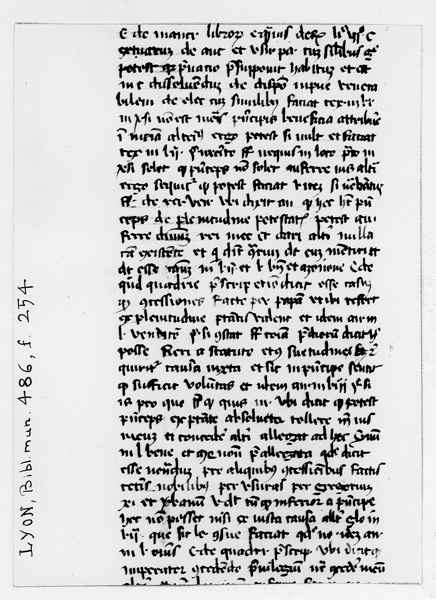}%
            \includegraphics[height=5cm]{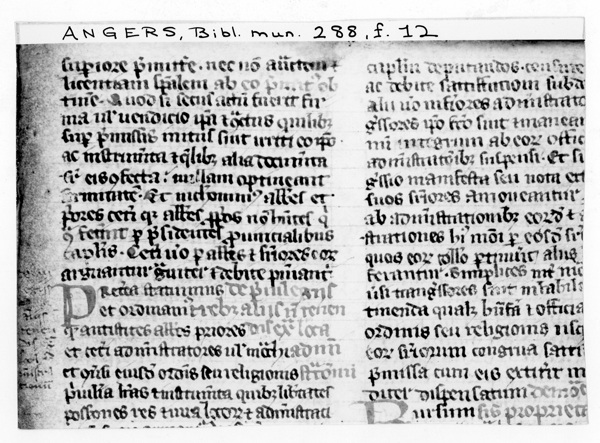}%
            \includegraphics[height=5cm]{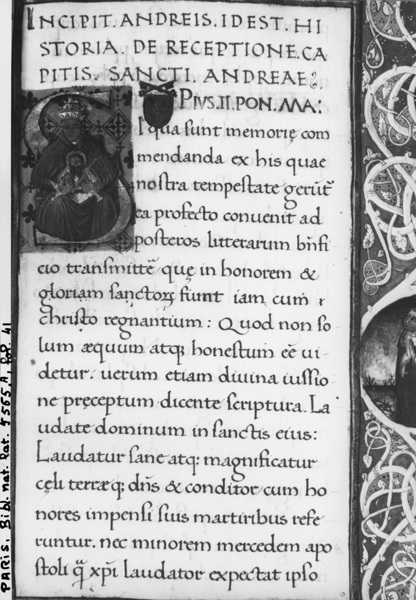}%
        }
        \resizebox{\linewidth}{!}{%
            \includegraphics[height=5cm]{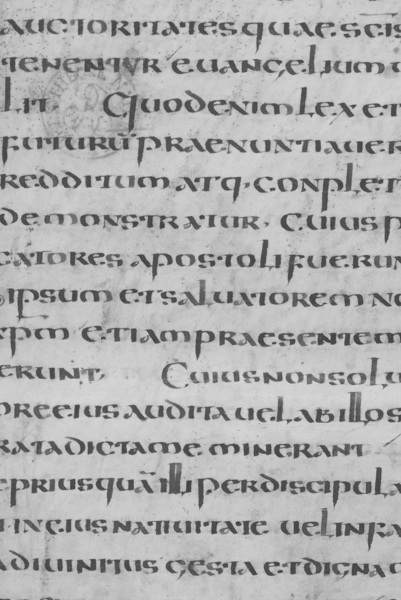}%
            \includegraphics[height=5cm]{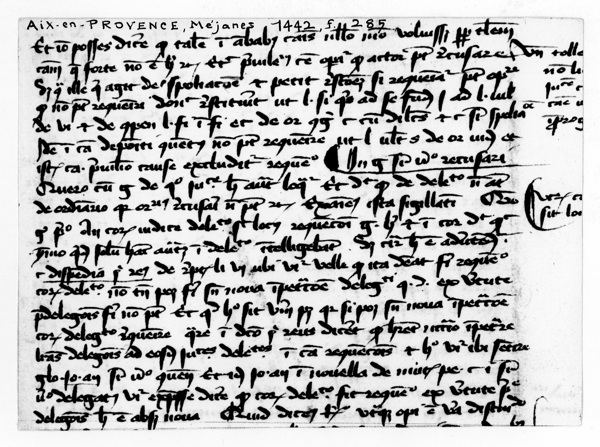}%
            \includegraphics[height=5cm]{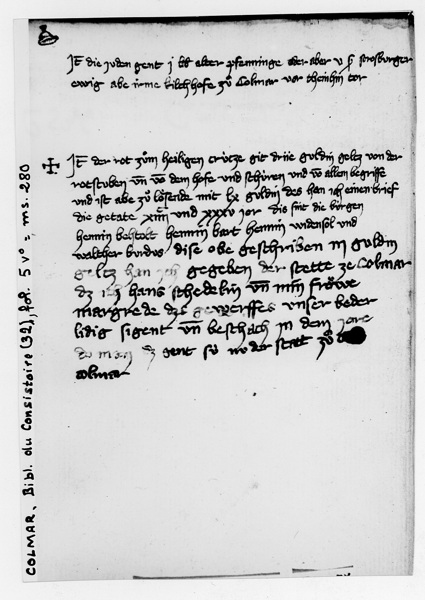}%
        }
        \caption{CLaMM}
    \end{subfigure}
    \caption{Example images from the datasets included in our benchmark. For EuroSAT, we only show the RGB bands.}
    \label{fig:dataset-examples}
\end{figure*}

\section{Experimental setup}
\label{sec:setup}

In this section, we give an overview of the experimental pipeline that we followed for a fair evaluation of the aforementioned methods on the five datasets that constitute our benchmark.

\begin{table*}[t]
    \centering
    \renewcommand{\arraystretch}{1.3}
    \setlength{\tabcolsep}{10pt}
    \rowcolors{2}{LightGray}{white}
    \begin{tabular}{l ccccc }
        \toprule
        Hyper-Parameter &    ciFAIR-10    &      CUB       &   ISIC 2018      &   EuroSAT    &     CLaMM\\
        \midrule
        Learning Rate   & \multicolumn{5}{c}{\texttt{loguniform}(1e-4, 0.1)}  \\
        Weight Decay    & \multicolumn{5}{c}{\texttt{loguniform}(1e-5, 0.1)} \\
        Batch Size      &         \{10, 25, 50\}  &   \{8, 16, 32\}  &  \{8, 16, 32\}  &   \{10, 25, 50\}   &   \{8, 16, 32\}\\
        Epochs          &   500  &   200  & 500  & 500  & 500 \\
        HPO Trials      & 250  &   100  &  100   &  250  & 100\\
        Grace Period    &     50  &   10  &  25  &  25  & 25\\
        \bottomrule
    \end{tabular}
    \caption{Summary of hyper-parameters searched/used with ASHA \cite{li2020system}. Method specific hyper-parameters were included in the search space but not included in this table due to space limitations. An epoch number in parentheses means that a higher number of epochs was used for the final training than for the hyper-parameter optimization.}
    \label{tab:hpo_params}
\end{table*}

\subsection{Evaluation metrics}
In our benchmark, we evaluate each method on each dataset with the widely used balanced classification accuracy.
This metric is defined as the average per-class accuracy, \ie, the average of the diagonal in the row-normalized confusion matrix.
We turned our attention toward this metric since some datasets in our benchmark do not have balanced test sets.
In any case, for balanced test sets, the balanced accuracy equals the standard classification accuracy.

Since our benchmark contains multiple datasets it is hard to directly make a comparison between two methods without computing an overall ranking.
Therefore, for each method, we also compute the average balanced accuracy across all datasets to provide a simple and intuitive way to rank methods.
Additionally, in this manner, future methods will be easily comparable with those already evaluated.

\subsection{Data pre-processing and augmentation} 
All input images were normalized by subtracting the channel-wise mean and dividing by the standard deviation computed on the \textit{trainval} splits.
We applied standard data augmentation policies with slightly varying configurations, adapted to the specific characteristics of each dataset and problem domain.
Note that none of the currently re-evaluated methods in our benchmark had as original contribution a specialized data augmentation technique.
Nothing prevents the use of a data-augmentation-based method from partaking in the benchmark.

For datasets with a small, fixed image resolution, \ie, ciFAIR-10 and EuroSAT, we perform random shifting by 12.5\% of the image size and horizontal flipping in 50\% of the cases.
For all other datasets, we apply scale augmentation using the \texttt{RandomResizedCrop} transform from PyTorch\footnote{\url{https://pytorch.org/vision/stable/transforms.html\#torchvision.transforms.RandomResizedCrop}} as follows:
A crop with a random aspect ratio drawn from $[\frac{3}{4}, \frac{4}{3}]$ and an area between $A_\mathrm{min}$ and 100\% of the original image area is extracted from the image and then resized to $224 \times 224$ pixels.
The minimum fraction $A_\mathrm{min}$ of the area was determined based on preliminary experiments to ensure that a sufficient part of the image remains visible.
It therefore varies depending on the dataset: we use $A_\mathrm{min} = 20\%$ for CLaMM and $A_\mathrm{min} = 40\%$ for CUB and ISIC 2018.

For ISIC 2018 and EuroSAT, we furthermore perform random vertical flipping in addition to horizontal flipping, since these datasets are completely rotation-invariant and vertical reflection augments the training sets without drifting them away from the test distributions.
On CLaMM, in contrast, we do not perform any flipping, since handwritten scripts are not invariant even against horizontal flipping.

\subsection{Architecture and optimizer}
To perform a fair comparison, we use the same backbone CNN architecture for all methods.
For ciFAIR-10, we employ a Wide Residual Network (WRN) \cite{zagoruyko2016wrn}, precisely WRN-16-8, which is widely used in the existing literature for data-efficient classification on CIFAR.
For all other cases, the popular and well-established ResNet-50 (RN50) architecture \cite{he2016deep} is used.
Note that we made changes to the architecture when that was an original contribution of the paper, but all those changes were applied to the selected base architecture.
Due to the high popularity of residual networks, the majority of the selected approaches were originally tested with a RN/WRN backbone.
This fact allowed us to perform a straightforward porting of the network setup, when necessary.

We furthermore employ a common optimizer and training schedule across all methods and datasets to avoid any kind of optimization bias.
We use standard stochastic gradient descent (SGD) with a momentum of 0.9, weight decay, and a cosine annealing learning rate schedule \cite{loshchilov2016sgdr}, which reduces the learning rate smoothly during the training process.
The initial learning rate and the weight decay factor are optimized for each method and dataset individually, together with any method-specific hyper-parameters as detailed in the next subsection.
The total number of training epochs for each dataset was chosen according to preliminary experiments.

\subsection{Hyper-parameter optimization}
\label{subsec:hpo}

Careful tuning of hyper-parameters, as one would do in practice, is crucial and can have a considerable impact on the final performance \cite{bischl2021hpo} (empirical proofs in  \cref{sec:results}).

For our benchmark, we hence first tune the hyper-parameters of each method on each individual dataset using the training and validation splits, which are disjoint from the test sets used for final performance evaluation (see \cref{sec:datasets}).
Since for each dataset we have three different sets of training splits, we perform three hyper-parameter optimization runs.
Only on CUB, we perform the three searches on the unique available training split.
For any method, we tune the initial learning rate and weight decay, sampled from a log-uniform space, as well as the batch size, chosen from a pre-defined set.
Details about the search space are provided in \cref{tab:hpo_params}.
In addition to these general hyper-parameters, any method-specific hyper-parameters are tuned as well simultaneously, considering the boundaries used in the original paper, if applicable, or lower and upper bounds estimated by ourselves.

For selecting hyper-parameter configurations to be tested and scheduling experiments, we employ Asynchronous HyperBand with Successive Halving (ASHA) \cite{li2020system} as implemented in the Ray library\footnote{\url{https://docs.ray.io/en/master/tune/}}.
This search algorithm exploits parallelism and aggressive early-stopping to tackle large-scale hyper-parameter optimization problems.
Trials are evaluated and stopped based on their accuracy on the validation split.

Two main parameters need to be configured for the ASHA algorithm: the number of trials and the grace period.
The former controls the number of hyper-parameter configurations tried in total while the latter the minimum time after which a trial can be stopped.
Since the number of trials corresponds to the time budget available for HPO, we choose larger values for smaller datasets, where training is faster.
The grace period, on the other hand, should be large enough to allow for a sufficient number of training iterations before comparing trials.
Therefore, we choose larger grace periods for smaller datasets, where a single epoch comprises fewer training iterations.
The exact values for each dataset as well as the total number of training epochs can be found in \cref{tab:hpo_params}.
These values were determined based on preliminary experiments with the cross-entropy baseline.

\subsection{Final training and evaluation}
After having completed HPO for each split using the procedure described above, we train the classifiers with the three determined configurations on the \emph{trainval\{i\}} splits and evaluate the balanced classification accuracy on the test splits.
To account for the effect of random initialization, this training is repeated ten times. 
Therefore, the final performance is the balanced average accuracy over 30 repetitions with each set of 10 repetitions initialized with the parameters found through HPO runs on the three sets of splits.

\begin{table*}[h]
    \centering
    \renewcommand{\arraystretch}{1.3}
    \setlength{\tabcolsep}{7pt}
    \rowcolors{0}{white}{LightGray}
    \resizebox{\linewidth}{!}{
    \begin{tabular}{lcccccc}
        \toprule
          Method    &   ciFAIR-10    &      CUB       &   ISIC 2018    &    EuroSAT     &     CLaMM      &      Average      \\
        \midrule
        Cross-Entropy Loss (untuned baseline)     &         46.52  &         53.24  &         57.33  &         83.48  &         48.12  &         57.74  \\
        Cross-Entropy Loss (baseline)        &         55.18  &         70.79  &         64.49  &         90.58  &         70.15  &         70.24  \\
        Deep Hybrid Networks (\textcolor{red}{$\bullet$}) \cite{oyallon2017scaling,oyallon2018scattering}  &         53.84  &         55.37  &         62.06  &         88.77  &         63.75  &         64.76  \\
        OL{\'E} (\textcolor{applegreen}{$\bullet$}) \cite{lezama2018ole}         &         55.19  &         66.55  &         62.80  &         90.29  &         74.28  &         69.82  \\
        Grad-\(\ell_{2}\) Penalty (\textcolor{applegreen}{$\bullet$}) \cite{bietti2019kernel}      &         51.90  &         51.94  &         60.21  &         81.50  &         65.10  &         62.13  \\
        Cosine Loss (\textcolor{applegreen}{$\bullet$}) \cite{barz2020deep}      &         52.39  &         66.94  &         62.42  &         88.53  &         68.89  &         67.83  \\
        Cosine + Cross-Entropy Loss (\textcolor{applegreen}{$\bullet$}) \cite{barz2020deep} &         52.77  &         70.43  &         63.17  &         89.65  &         70.64  &         69.33  \\
        Harmonic Networks (\textcolor{red}{$\bullet$}) \cite{ulicny2019harmonicnet,ulicny2019harmonic}    & \textbf{58.00} & \textbf{73.07} & \textbf{69.70} & \textbf{91.98} & \textbf{77.25} & \textbf{74.00} \\
        Full Convolution (\textcolor{red}{$\bullet$}) \cite{kayhan2020translation}       &         54.64  &         63.74  &         57.34  &         89.47  &         69.06  &         66.85  \\
        Dual Selective Kernel Networks (\textcolor{red}{$\bullet$}) \cite{sun2020visual}         &         53.84  &         69.75  &         63.41  &         91.09  &         65.43  &         68.70  \\
        Distilling Visual Priors (\textcolor{blue}{$\bullet$}\textcolor{applegreen}{$\bullet$}) \cite{zhao2020distilling}     & \textit{57.80} &         70.81  &         62.39  &         88.96  &         69.07  &         69.81  \\
        Auxiliary Learning (\textcolor{applegreen}{$\bullet$}) \cite{navon2021auxiliary}   &         51.84  &         43.57  &         61.70  &         80.92  &         60.24  &         59.65  \\
        T-vMF Similarity (\textcolor{applegreen}{$\bullet$}) \cite{kobayashi2021t}        &         56.75  &         68.19  &         64.60  &         88.50  &         69.33  &         69.47  \\
        \bottomrule
    \end{tabular}
    }

    \caption{Average balanced classification accuracy in \% over 30 repetitions for each task and across all tasks. 
    The best value per dataset is highlighted in bold font. Numbers in italic font indicate that the result is not significantly worse than the best one on a significance level of 5\%.
    Colored dots represent the taxonomic classes to which the approaches belong, \ie, \emph{architecture} (\textcolor{red}{$\bullet$}), \emph{cost function} (\textcolor{applegreen}{$\bullet$}), and \emph{warm-starting} (\textcolor{blue}{$\bullet$}).}
    \label{tab:results}
\end{table*}

\subsection{Method-specific implementation details}


Two methods required individual modifications to the general training and evaluation pipeline described above, which we describe in the following.

\subsubsection{Grad-\(\ell_{2}\) Penalty}
We disabled weight decay because this method is an alternative regularizer and considered as mutually exclusive with weight decay in the original paper \cite{bietti2019kernel}.
Moreover, in contrast to the original implementation, we enabled the use of batch normalization since, without this component, we obtained extremely low results in preliminary experiments.

\subsubsection{Distilling Visual Priors}

DVP \cite{zhao2020distilling} required several adaptations due to its two-step training process.

First, task-agnostic features are learned using self-supervision.
Then, the resulting model is used as a teacher for a student model trained for classification.
Thus, not only the training of the final classifier needs tuned hyper-parameters but the pre-training step as well.
For evaluating the quality of the pre-trained models during HPO, we attach a linear classification head on top of the learned representations but do not back-propagate through it.
The contrastive pre-training criterion furthermore requires larger batch sizes than we usually use and longer training schedules.
We hence select batch sizes from \{64, 128, 256\} and increase the number of epochs until the accuracy of the additional classification head converges.
This resulted in a training schedule of 2,000 epochs for ciFAIR-10, 6,400 epochs for CUB, and 16,000 epochs for ISIC 2018, CLaMM, and EuroSAT.
These numbers are one to two orders of magnitude larger than our usual training durations used for all other methods.

After we found hyper-parameters for the self-supervised pre-training, we trained a single model on the training split, which served as a basis for the subsequent HPO for distilling the learned knowledge into the student classifier.
For this step, we used the same batch sizes and numbers of epochs as usual.
Finally, we trained 30 self-supervised models, on the combined training and validation data and subsequently used each of them as a teacher for 30 other student models performing the classification task.
As for the other methods, HPO and the final training of each group of 10 are performed on a different set of dataset splits.

\section{Results}
\label{sec:results}

In the following, we first present the main results of our benchmark in \cref{subsec:benchmark}, followed by comparisons in terms of training speed and memory requirements (\cref{subsec:computation}), and between taxonomic classes (\cref{subsec:comptaxonomy}).
We then show an analysis concerning an additional high-resolution fine-tuning step in \cref{subsec:highres-finetune}.
In \cref{subsec:transfer}, we compare benchmarked methods with transfer learning.
Next, in \cref{subsec:underbaselines}, we provide evidence that published baselines are underperforming.
The importance of hyper-parameter optimization is discussed in \cref{subsec:gridsearch}.
Finally, in \cref{subsec:optimalhparams} we provide additional insights on the tuned hyper-parameters of the cross-entropy baseline.

\subsection{Benchmark for Image Classification with Small Datasets}
\label{subsec:benchmark}

\Cref{tab:results} presents the average balanced classification accuracy over 30 repetitions for all methods and datasets.
We performed Welch's t-test to assess the significance of the advantage of the best method per dataset in comparison to all others.
All results but one are significantly worse on a level of 5\% than the best method on the respective task.

Harmonic Networks \cite{ulicny2019harmonicnet,ulicny2019harmonic} clearly win the benchmark by providing top performance on all datasets.
However, an even more interesting finding of this study is that the default cross-entropy loss is highly competitive when tuned carefully.
In terms of average balanced accuracy across all datasets, the baseline scores 70.24\%, which is only clearly below the accuracy scored by Harmonic Networks (74\%) but superior than or on par with the results of the remaining methods.
To better characterize the impact of hyper-parameter optimization (HPO) on the baseline, we also perform experiments with the cross-entropy loss and default hyper-parameters (untuned baseline in \cref{tab:results}).
By default hyper-parameters, we mean the learning rate and weight decay that are usually employed in data-rich scenarios, \ie, 0.1 for the first and \(10^{-4}\) for the latter.
Without HPO there is a substantial degradation of the baseline performance resulting in a very low average balanced accuracy across all tasks.
The untuned baseline only scores \(57.74\%\) which is \textapprox13 percentage points below the tuned baseline and clearly outperformed by all other specialized methods.

A relatively large group of state-of-the-art approaches, \ie, OL{\'E} \cite{lezama2018ole}, Cosine + Cross-Entropy Loss \cite{barz2020deep}, Dual Selective Kernel Networks \cite{sun2020visual}, Distilling Visual Priors \cite{zhao2020distilling}, and T-vMF Similarity \cite{kobayashi2021t}, obtains an overall recognition performance comparable to the one of the baseline.
However, the finding that the vast majority of recent methods for image classification with small datasets does not exceed the performance of the baseline is sobering.
We attribute this to the fact that the importance of hyper-parameter optimization is immensely underestimated, resulting in misleading comparisons of novel approaches with weak and underperforming baselines.
We will investigate this hypothesis further in later sub-sections.

\begin{figure}[t]
\centering
    \resizebox{\linewidth}{!}{
            \input{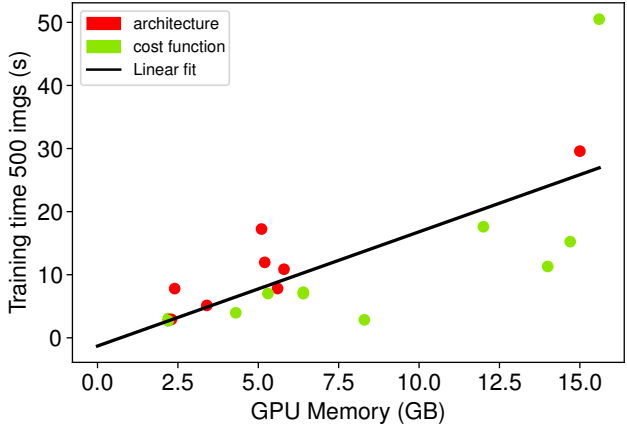}}
    \caption{Comparison between \emph{architecture} (\textcolor{red}{$\bullet$}) and \emph{cost function} (\textcolor{applegreen}{$\bullet$}) methods in terms of GPU memory usage and training time (500 images).
    Training times are derived from the training speeds shown in \cref{tab:computation}.}
    \label{fig:timevsmemory}
\end{figure}

\subsection{Time and memory requirements}
\label{subsec:computation}

\begin{table}[t]
    \centering
    \renewcommand{\arraystretch}{1.3}
    \setlength{\tabcolsep}{9pt}
    \rowcolors{3}{white}{LightGray}
    \resizebox{\linewidth}{!}{
    \begin{tabular}{l cc cc}
        \toprule
        & \multicolumn{2}{c}{Images/sec ($\uparrow$)} & \multicolumn{2}{c}{
        {GPU Mem. ($\downarrow$)}} \\
        \cmidrule(lr){2-3}\cmidrule(l){4-5}\rowcolor{white}
        \multirow{-2}{*}{Method} & 224 & 448 & 224 & 448 \\
        \midrule
        Cross-Entropy Loss (baseline)         & \textbf{189.0} & \textbf{72.2} &  \textbf{2.2} &  6.4 \\
        Deep Hybrid Networks (\textcolor{red}{$\bullet$})           &  64.1 & 29.0 &  2.4 &  \textbf{5.1} \\
        OL{\'E} (\textcolor{applegreen}{$\bullet$})                        & 163.7 & 69.2 &  \textbf{2.2} &  6.4 \\
        Grad-\(\ell_{2}\) Penalty (\textcolor{applegreen}{$\bullet$})      &  28.4 &  9.9 & 12.0 & 15.6 \\
        Cosine + Cross-Entropy Loss (\textcolor{applegreen}{$\bullet$})    & 183.8 & 71.4 &  \textbf{2.2} &  6.4 \\
        Harmonic Networks (\textcolor{red}{$\bullet$})              & 171.0 & 64.0 &  2.3 &  5.6 \\
        Full Convolution (\textcolor{red}{$\bullet$})               &  97.3 & 46.0 &  3.4 &  5.8 \\
        Dual Selective Kernel Networks (\textcolor{red}{$\bullet$}) &  41.8 & 16.9 &  5.2 & 15.0 \\
        Distilling Visual Priors (\textcolor{blue}{$\bullet$}\textcolor{applegreen}{$\bullet$}) & & & & \\
        \rowcolor{white}
        \qquad Pre-Training            & 175.0 &   -- &  8.3 &   -- \\
        \qquad Fine-Tuning             & 125.9 & 44.2 &  4.3 & 14.0 \\
        Auxiliary Learning (\textcolor{applegreen}{$\bullet$})             &  32.8 &   -- & 14.7 & 30.3 \\
        T-vMF Similarity (\textcolor{applegreen}{$\bullet$})               & 184.8 & 71.3 &  \textbf{2.2} &  5.3 \\
        \bottomrule
    \end{tabular}
    }
    \caption{Comparison between methods in terms of training speed (images/second) and GPU memory requirements (GB) for input resolutions of $224 \times 224$ and $448 \times 448$ on an NVIDIA V100.
    The best value per column is reported in bold font.
    Colored dots represent the taxonomic classes to which the approaches belong, \ie, \emph{architecture} (\textcolor{red}{$\bullet$}), \emph{cost function} (\textcolor{applegreen}{$\bullet$}), and \emph{warm-starting} (\textcolor{blue}{$\bullet$}).}
    \label{tab:computation}
\end{table}

\begin{figure}[t]
\centering
    \resizebox{\linewidth}{!}{
            \input{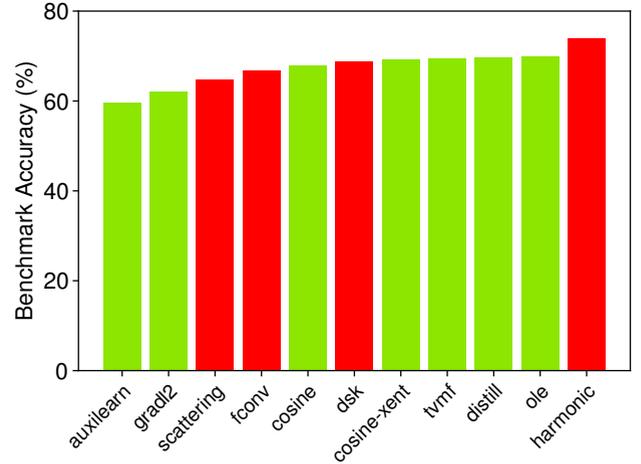}}
    \caption{Comparison between \emph{architecture} (\textcolor{red}{$\bullet$}) and \emph{cost function} (\textcolor{applegreen}{$\bullet$}) methods in terms of recognition performance.
    We order methods along the x-axis from the least to the best performing one.}
    \label{fig:acc_taxonomy}
\end{figure}

In the previous sub-section, we compared the benchmarked methods in terms of accuracy.
Further important factors for choosing a method from a practical point of view are the training speed and memory requirements.
We measured these two metrics on an Nvidia V100 GPU with 16 GB of memory using PyTorch 1.7.

For measuring training throughput, we run each training method for 10 epochs on the CUB dataset and count the number of images processed per second during the last 5 epochs.
For a fair comparison, we used a constant batch size of 8 for all methods, which was the most common batch size found by HPO.
The only exception is the self-supervised pre-training step of Distilling Visual Priors, which depends on sufficiently large batch sizes.
For this method, we used a batch size of 64, as determined by HPO.

In \cref{tab:computation}, we report training throughput and GPU memory requirements for input resolutions of $224 \times 224$ and $448 \times 448$.
We also report the latter resolution to gain additional insights regarding how computational requirements scale with larger inputs.
As we will see in \cref{subsec:highres-finetune}, some tasks benefit from larger resolutions.

Five among the top methods concerning recognition accuracy, namely Harmonic Networks, the baseline, OL{\'E}, Cosine + Cross-Entropy Loss, and T-vMF Similarity, are also among the fastest and most memory-efficient ones.
The slowest and most memory-consuming methods are Auxiliary Learning and Grad-\(\ell_{2}\) Penalty, which take 5-7 times longer to train than the baseline and consume 5-7 times more memory.
In addition, they are the worst-performing methods in terms of classification accuracy.
For Auxiliary Learning, the memory of the V100 was insufficient for the higher resolution of $448 \times 448$, wherefore we measured the memory consumption in this case on an A100 GPU with 40 GB of memory.
Due to the different compute hardware, we do not report the training throughput in this case.

\subsection{Comparison between taxonomic classes}
\label{subsec:comptaxonomy}

In this paragraph, we briefly discuss the performance difference among the two largest taxonomic groups evaluated on our benchmark, \ie, the \emph{architecture} (\textcolor{red}{$\bullet$}) and \emph{cost function} (\textcolor{applegreen}{$\bullet$}) classes.

First, we compare these two classes considering the recognition performance over the benchmark.
In \cref{fig:acc_taxonomy}, we show the results of each method on the full benchmark in ascending order.
We notice that colored bars follow an alternating pattern, i.e., the worst two methods belong to the \emph{cost function} class, then \emph{architecture}, and so on. 
Hence, there is no clear winner considering this evaluation metric.
Harmonic Networks \cite{ulicny2019harmonicnet,ulicny2019harmonic}, which belong to the \emph{architecture} class, however, remain the winners of our benchmark.

Second, we discuss the differences concerning computational requirements and show the comparison in \cref{fig:timevsmemory}.
We plot the training time needed to process 500 images vs the GPU memory usage.
Training times are derived from the speeds shown in \cref{tab:computation}.
We opted for 500 images because they well represent the time to perform a single epoch on a small dataset (e.g., the size of our ciFAIR-10 training splits).
We notice that the two taxonomic classes have similar computational requirements but lie on opposite sides of a linear fit.
The majority of red dots are above, while the green ones are below.
Hence, \emph{architecture} methods seem to be slower to train while \emph{cost function} ones to require more GPU memory. 
This is reasonable since the introduction of modified modules in the architecture may induce more complex processing and hence lower training speeds.
Modified losses may instead require gradient updates that consume more memory. 

\subsection{High-resolution fine-tuning}
\label{subsec:highres-finetune}

In the field of fine-grained visual recognition, it is common practice to increase the input image resolution from $224 \times 224$ pixels usually used for pre-training to $448 \times 448$ pixels \cite{cui2018large,lin2015bilinear}.
Therefore, in these experiments, we test such an additional step for the CUB and CLaMM datasets.
Initially, we also experimented with high-resolution fine-tuning for ISIC 2018 but did not observe any substantial advantage.
We run the test on ten models that have been trained on the training set with the standard resolution.
Such networks are fine-tuned with the double input image size of $448 \times 448$, which is, consequently, also used for evaluation.
For this high-resolution fine-tuning step, we use the same hyper-parameters and number of epochs as for the initial standard-resolution training.

OL{\'E} \cite{lezama2018ole}, Full Convolution \cite{kayhan2020translation},  and T-vMF Similarity \cite{kobayashi2021t} benefit the most from the high-resolution fine-tuning step on CUB and CLaMM, which improves the balanced accuracy by 18\%-23\% on CLaMM and 15\%-18\% on CUB for these methods.
Deep Hybrid Networks \cite{oyallon2017scaling,oyallon2018scattering}, in contrast, take the least advantage from the higher resolution.
On CLaMM, they only gain \textapprox3 percentage points.
The baseline is in-between with an improvement of 8\% on CLaMM and 13\% on CUB.
For AuxiLearn \cite{navon2021auxiliary}, we omitted the high-resolution fine-tuning step since the computational cost for this method becomes too high with the increase of resolution (see \cref{subsec:computation}).
Training would have taken several months to complete, which is by no means advantageous. 

We believe that this analysis turns out to be useful for practitioners facing problems with data scarcity in fine-grained scopes.
High-resolution fine-tuning can be performed to raise the recognition performance of the classifier through a relatively straightforward additional training step.

\begin{table}[t]
    \renewcommand{\arraystretch}{1.3}
    \setlength{\tabcolsep}{6pt}
    \rowcolors{0}{white}{LightGray}
    \resizebox{\linewidth}{!}{
    \begin{tabular}{l cccc}
        \toprule
        Method &  \multicolumn{2}{c}{CUB}     &    \multicolumn{2}{c}{CLaMM}\\
        \rowcolor{white} \(448 \times 448\) fine-tuning &       \ding{55}      &       \ding{51}      &       \ding{55}      &       \ding{51} \\
        \midrule
        Cross-Entropy Loss (baseline)                                              &    71.44  & 80.57 & 75.34 &         81.53 \\
        Deep Hybrid Networks (\textcolor{red}{$\bullet$}) &         52.54  &         58.57  &         65.74  &         68.00\\
        OL{\'E} (\textcolor{applegreen}{$\bullet$})                                          &         63.32  &         73.33  &         71.42  & 83.72\\
        Grad-\(\ell_{2}\) Penalty (\textcolor{applegreen}{$\bullet$})                     &         51.94  &         61.04  &         65.10  &         76.95 \\
        Cosine Loss (\textcolor{applegreen}{$\bullet$})                                       &         66.94  &         73.66  &         68.89  &         79.03\\
        Cosine + Cross-Entropy Loss (\textcolor{applegreen}{$\bullet$})                       &         70.80  &         78.08  &         69.29  &         79.90\\
        Harmonic Networks (\textcolor{red}{$\bullet$})      & 72.26 & 80.49 & 74.59 & 83.41\\
        Full Convolution (\textcolor{red}{$\bullet$})                         &        64.90  &         76.37  &         63.33  &         77.91\\
        Dual Selective Kernel Networks (\textcolor{red}{$\bullet$})                   &        71.02  &         78.84  &         61.51  &         63.45\\
        Distilling Visual Priors (\textcolor{blue}{$\bullet$}\textcolor{applegreen}{$\bullet$})                    &71.27  &         77.59  &         67.89  &         75.92\\
        Auxiliary Learning (\textcolor{applegreen}{$\bullet$})                          &         59.68  &           ---  &         43.61  &           --- \\
        T-vMF Similarity (\textcolor{applegreen}{$\bullet$})                               &         67.43  &         77.26  &         66.40  &         78.16\\
        \bottomrule
    \end{tabular}
    }

    \caption{Impact of high-resolution fine-tuning step on two fine-grained datasets, \ie, CUB and CLaMM. We report the average balanced classification accuracy in \% over 10 repetitions.
    Colored dots represent the taxonomic classes to which the approaches belong, \ie, \emph{architecture} (\textcolor{red}{$\bullet$}), \emph{cost function} (\textcolor{applegreen}{$\bullet$}), and \emph{warm-starting} (\textcolor{blue}{$\bullet$}).}
    \label{tab:highres-finetune}
\end{table}

\begin{table}[t]
    \centering
    \renewcommand{\arraystretch}{1.3}
    \setlength{\tabcolsep}{8pt}
    \rowcolors{2}{LightGray}{white}
    \resizebox{\linewidth}{!}{
    \begin{tabular}{lccc}
        \toprule
                              & ISIC 2018 & EuroSAT & CLaMM \\
        \midrule
        Best From Scratch     &  69.70  &   91.98   & 77.25 \\
        ImageNet Pre-Training &  75.92  &   94.22   & 75.21 \\
        Absolute Gap          &  \textcolor{red}{-6.22} &  \textcolor{red}{-2.24} & \textcolor{darkpastelgreen}{+2.04}\\
        \bottomrule
    \end{tabular}
    }
    \caption{Comparison between the best performance when training from scratch according to \cref{tab:results} and fine-tuning from weights pre-trained on ImageNet-1k. Numbers are the average balanced classification accuracy in \% over 30 repetitions.}
    \label{tab:finetuning}
\end{table}

\begin{table*}[t]
    \centering
    \renewcommand{\arraystretch}{1.3}
    \setlength{\tabcolsep}{5pt}
    \resizebox{\linewidth}{!}{
    \begin{tabular}{ccccccccccc}
        \toprule
        \multicolumn{5}{c}{Cross-Entropy Loss (baseline)} &  & \multicolumn{5}{c}{Other Methods} \\ 
        \cmidrule{1-5} \cmidrule{7-11}
        Publication  & Dataset   &  Network & Accuracy  &  Abs. Gap & & Method & Dataset & Network & Accuracy&  Abs. Gap \\
        \cmidrule{1-5} \cmidrule{7-11}
        
         \cellcolor{LightGray}\cite{oyallon2017scaling}  & \cellcolor{LightGray}CIFAR-10  & \cellcolor{LightGray}WRN-16-8  &  \cellcolor{LightGray}46.50 & \cellcolor{LightGray}  & & {DHN \cite{oyallon2017scaling}}\cellcolor{LightGray} & CIFAR-10 \cellcolor{LightGray} & WRN-16-8 \cellcolor{LightGray} & 54.70 \cellcolor{LightGray} & \cellcolor{LightGray}\\
         
         \cellcolor{LightGray}Ours  & \cellcolor{LightGray}ciFAIR-10  &  \cellcolor{LightGray}WRN-16-8  &  \cellcolor{LightGray}55.18  & \cellcolor{LightGray} \textcolor{darkpastelgreen}{+8.68} & & DHN (Ours)  \cellcolor{LightGray}  & ciFAIR-10 \cellcolor{LightGray}& WRN-16-8 \cellcolor{LightGray}&  53.84 \cellcolor{LightGray} & \cellcolor{LightGray}\textcolor{red}{-0.86}\\

         \cite{ulicny2019harmonic} & CIFAR-10 &  WRN-16-8  & 52.20  &  & & HN \cite{ulicny2019harmonic}  & CIFAR-10   & WRN-16-8   & 58.40  & \\
         
         Ours  & ciFAIR-10  &  WRN-16-8  &  55.18  & \textcolor{darkpastelgreen}{+2.98} & & HN (Ours)   & ciFAIR-10 & WRN-16-8 &  58.00  & \textcolor{red}{-0.40}\\
         
        
        
         \cellcolor{LightGray}\cite{navon2021auxiliary} & \cellcolor{LightGray}CUB & \cellcolor{LightGray}RN18 & \cellcolor{LightGray}37.20  & \cellcolor{LightGray}& & \cellcolor{LightGray}AuxiLearn \cellcolor{LightGray}\cite{navon2021auxiliary} & \cellcolor{LightGray}CUB & \cellcolor{LightGray}RN18 &  \cellcolor{LightGray}44.50  & \cellcolor{LightGray}\\
         \cellcolor{LightGray}Ours & \cellcolor{LightGray}CUB & \cellcolor{LightGray}RN18 & \cellcolor{LightGray}65.80  & \cellcolor{LightGray}\textcolor{darkpastelgreen}{+28.60}& & \cellcolor{LightGray}AuxiLearn (Ours) & \cellcolor{LightGray}CUB & \cellcolor{LightGray}RN18 &  \cellcolor{LightGray}56.00  & \cellcolor{LightGray}\textcolor{darkpastelgreen}{+11.50} \\
         
          \cite{barz2020deep}& CUB & RN50 & 51.92 & & & Cosine Loss \cite{barz2020deep} & CUB & RN50 &  67.60 &   \\
          
          Ours & CUB & RN50 & 70.79  & \textcolor{darkpastelgreen}{+18.87}& & Cosine Loss (Ours) & CUB & RN50 &  66.94  & \textcolor{red}{-0.66}\\
        
        \bottomrule
    \end{tabular}
    }
    \caption{Summary of published/our results of the cross-entropy loss (left) and other methods (right) on similar setups. }
    \label{tab:baseline-comparison}
\end{table*}

\subsection{Strengths and limits of transfer learning}
\label{subsec:transfer}

In scenarios where it is possible, so-called \emph{transfer learning} by pre-training on a large related dataset and fine-tuning on the target data is a popular technique.
It does not qualify for our benchmark due to the use of external data, but a comparison with this approach allows us to understand its benefits and limitations.
ImageNet pre-training particularly benefits down-stream tasks whose labels are well-represented in ImageNet (\eg, CUB) \cite{kornblith2019better}.
Yet, we show that the outcome changes as the target domain moves away from the one of natural images.

In this set of experiments, we fine-tune a ResNet-50 pre-trained on ImageNet-1k on the datasets of our benchmark which do not contain natural images, \ie, ISIC 2018, EuroSAT, and CLaMM using the standard cross-entropy loss and compare it with the best small-data method in \cref{tab:finetuning}.
The hyper-parameters for the fine-tuning step are tuned in the same manner as for our benchmark (see \cref{subsec:hpo}).
For EuroSAT, which is a multispectral dataset, we have to restrict the fine-tuning to the R, G, and B channels to be feasible.

\emph{Transfer learning} has a clear advantage on the ISIC dataset. The latter, despite being from a different domain (medical), shares low-level textures and colors with natural images.
On EuroSAT, which comprises satellite images, we note a smaller accuracy difference (2.24 percent points).
Finally, on CLaMM (manuscript imagery) the domain shift is detrimental: training from scratch outperforms fine-tuning by 2 percent points.
Here, an additional factor that might play a role besides the domain shift is the data modality, since CLaMM contains only grayscale images.

We learn from this analysis that \emph{transfer learning} may only be applicable in data-deficient scenarios that share low-level features with natural images, but fail as the domain shift becomes more significant.

\begin{figure*}[t]
    \resizebox{\linewidth}{!}{\input{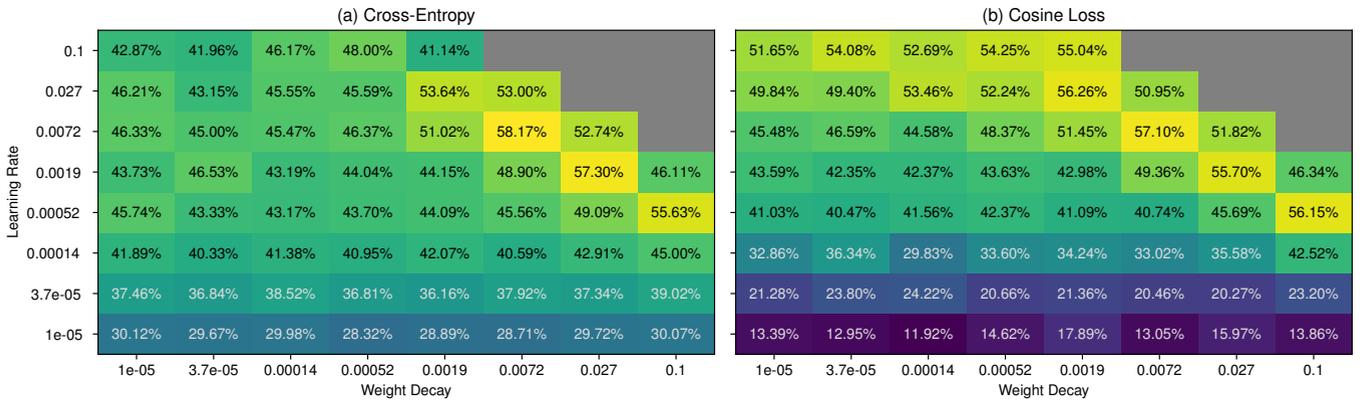}}%
    {\phantomsubcaption\label{subfig:gridsearch-xent}}%
    {\phantomsubcaption\label{subfig:gridsearch-cosine}}%
   \caption{Classification accuracy obtained with standard cross-entropy and cosine loss \cite{barz2020deep} on ciFAIR-10 with 1\% of the training data for different combinations of learning rate and weight decay. Gray configurations led to divergence.}
   \label{fig:gridsearch}
\end{figure*}

\subsection{Published baselines are underperforming}
\label{subsec:underbaselines}

We show further evidence of why tuning the hyper-parameters and not neglecting the baseline in a small-data setting is fundamental for performing a fair comparison between different methods.

We analyzed the original results reported for the methods considered in our study and compare those that share a similar setup with the performance of our re-evaluation.
Furthermore, we compare the performance of the baseline published in those works with ours.
Note that due to the lack of a standard benchmark and the common practice of randomly sub-sampling large datasets, we are unable to conduct a fair comparison with the same dataset splits, training procedure, etc.
Still, our benchmark shares the base dataset and network architecture with the selected cases.
Therefore, we believe that this analysis is suitable for supporting our point regarding the common practice of comparing tuned proposed methods with underperforming baselines.
The results of this analysis are shown in \cref{tab:baseline-comparison}.

Deep Hybrid Networks and Harmonic Networks were originally tested with a WRN-16-8 on CIFAR-10.
Full Convolution and Cosine Loss employed RN50 on CUB.
Since in the original Auxiliary Learning paper the authors trained a ResNet-18 (RN18) on CUB \cite{navon2021auxiliary}, we also perform an additional experiment with this smaller architecture to perform a fair comparison.
The training set of CIFAR-10 was comprised of \(50\) images per class.
Differently, experiments on CUB employed the full training set (\ie, 30 images per class).

Our baseline clearly outperforms the published baselines by large margins (\cref{tab:baseline-comparison}, left part).
More precisely, our models surpass the published ones by \textapprox 9, \textapprox 3, \textapprox 29, and \textapprox 19 percentage points on the CIFAR and CUB setups.
Recall also that the ciFAIR-10 test set is slightly harder than the CIFAR-10 one due to the removal of duplicates \cite{barz2020cifair}.

The picture looks different in the case of the proposed methods (\cref{tab:baseline-comparison}, right part), where the difference between ours and the original results is sharply less evident.
Our DHN and HN slightly underperform the original ones by \textapprox 1.0 and \textapprox 0.50 percentage points, respectively.
However, this was expected due to the higher difficulty of ciFAIR-10. 
On CUB, our re-evaluation of the cosine loss scores an average balanced accuracy of 66.94\% which is very close to the original 67.60\%.
Finally, our AuxiLearn model employing RN18 gains \textapprox 11 percentage points confirming once again that careful HPO can further boost the performance.

From this analysis, it seems clear that proposed methods are usually tuned to obtain an optimal or near-optimal result while baselines are trained with default hyper-parameters that have been found useful for large datasets but do not necessarily generalize to smaller ones.

\subsection{Importance of hyper-parameter optimization}
\label{subsec:gridsearch}

To further underpin the importance of hyper-parameter optimization, especially in a data-deficient setting, we perform a full grid search for combinations of learning rate and weight decay with a Wide ResNet architecture \cite{zagoruyko2016wrn} trained on as few as 1\% of the CIFAR-10 training data \cite{krizhevsky2009learning} and evaluated on the ciFAIR-10 test set \cite{barz2020cifair}.
We conduct this experiment with the standard cross-entropy loss and with the cosine loss \cite{barz2020deep}, which was proposed as a loss function with a regularizing effect for better performance on small datasets.

First, the results for standard cross-entropy shown in \cref{subfig:gridsearch-xent} illustrate that this baseline can substantially benefit from suitable HPO.
Typical default hyper-parameters such as a learning rate of 0.1 and weight decay of \num{1e-4} as used by \cite{bruintjes2021vipriors} would achieve \textapprox 46\% accuracy in this scenario, which is entire 12 percentage points below the optimal performance of \textapprox 58\%.

In comparison with the results for cosine loss shown in \cref{subfig:gridsearch-cosine}, we observe that the cosine loss is less sensitive to changes of the weight decay factor but more sensitive to the learning rate.
For the experiments in the original cosine-loss paper \cite{barz2020deep}, the authors did perform HPO for both methods but only took the learning rate into account while keeping the weight decay fixed to a small constant.
In this setting, the cosine loss can easily outperform cross-entropy because it has better chances with arbitrary weight decays.
For 6 out of the 8 weight decay values we tested, cosine loss achieves better performance than cross-entropy.
Only when both hyper-parameters are tuned can the cross-entropy baseline demonstrate its strength.

It is furthermore worth noting that the optimal weight decay in this data-deficient setting is rather large compared to usual defaults, which range between \num{1e-5} and \num{1e-4}.
Such small training datasets apparently require much stronger regularization.

Moreover, we observe that the best performing hyper-parameter combinations are close to an area of the search space that results in divergence of the training procedure.
This makes hyper-parameter optimization a particularly delicate endeavor.

\subsection{Tuned hyper-parameters}
\label{subsec:optimalhparams}

 \begin{table}[t]
     \setlength{\tabcolsep}{4pt}
     \renewcommand{\arraystretch}{1.3}
     \begin{tabularx}{\linewidth}{Xccc}
         \toprule
          Dataset   & Batch Size & Learning Rate & Weight Decay \\
          \midrule
          \rowcolor{LightGray}
          ciFAIR-10 &         10 & \num{4.55e-3} & \num{5.29e-3} \\
          CUB       &          8 & \num{2.44e-3} & \num{2.22e-3} \\
          \rowcolor{LightGray}
          ISIC 2018 &          8 & \num{0.69e-3} & \num{4.16e-2} \\
          EuroSAT   &         25 & \num{4.82e-3} & \num{6.31e-2} \\
          \rowcolor{LightGray}
          CLaMM     &          8 & \num{1.81e-3} & \num{1.81e-2} \\
          \bottomrule
     \end{tabularx}
     \caption{Hyper-parameters found with ASHA \cite{li2020system} for the cross-entropy baseline on one of the training splits.}
     \label{tab:hparams}
 \end{table}


For reproducibility, but also to gain further insights into hyper-parameter optimization for small datasets, we show one of the three hyper-parameter combinations found during our searches for the cross-entropy baseline in \cref{tab:hparams}.

We can observe that small batch sizes seem to be beneficial, despite the use of batch normalization.
While the learning rate exhibits a rather small range of values from \num{0.7e-3} to \num{7.4e-3} across datasets and spans only one order of magnitude, weight decay varies within a range of two orders of magnitude from \num{4.1e-4} to \num{1.8e-2}.

Furthermore, learning rate and weight decay appear to be negatively correlated.
Higher learning rates are usually accompanied by smaller weight decay factors.
The same correlation can be observed in \cref{fig:gridsearch}.

\section{Discussion}
\label{sec:discussion}

We designed a rigorous evaluation protocol for each method based on a common experimental setup in terms of base architecture, dataset splits, and optimization pipeline.
However, we acknowledge that our study is not inclusive with respect to all possible aspects.
In the following, we provide a list of the possible limitations of our benchmark along with explanations and arguments concerning each of them.

\subsection{Datasets}

An extensive focus on individual benchmark datasets and even certain dataset splits bear the risk of adapting methods specifically to the test sets of these few datasets.
To account for random variations caused, for instance, by data sub-sampling, we run our experiments on three independent dataset splits.
To ensure the generalization of the tested methods across domains, our benchmark transcends the common datasets, \eg, CIFAR, and incorporates four additional datasets with widely varying characteristics.
These additions augment the generality of our benchmark, yet keep a balance between the spectrum of covered domains and the overall computation time needed to evaluate a method.
However, given the "living" nature of our benchmark, we plan in the future to introduce domains spanning an even broader range of fields, data types, and applications to drive further progress toward small-sample learning methods.

\subsection{Base Architecture}

Concerning the base architecture employed, ResNet is not only a quite popular architecture in the image classification literature, but also in the image classification with small datasets community, as we saw from our literature analysis.
We employed this network class to remain consistent with previous literature but we would not exclude an architectural bias for ResNets a priori.
However, including multi-architecture evaluations in such a large benchmark like ours would incur high computational costs.

\subsection{Hyper-Parameter Search}

The strong performance of our baseline and the comparison with published baselines in \cref{tab:baseline-comparison} demonstrate the importance of thorough hyper-parameter optimization.
Concerning this aspect, our benchmark is fair in the sense that all methods had the same budget (in terms of the number of HPO trials) and we tuned all their hyper-parameters jointly.
However, comparing the grid-search results for the cosine loss on the ciFAIR-10 test set (see \cref{subfig:gridsearch-cosine}) with the respective performance reported in \cref{tab:results} exhibits a gap of about 5 percentage points.
Obviously, in this case, our HPO procedure did not find hyper-parameters on the validation sets that provide optimal performance on the test set.
We conjecture two main reasons that may have caused this failure: the search algorithm could not find the optimal solutions or the evaluation performance obtained on the validation sets does not directly translate into optimal performance on the test set.
Our correct practice of performing HPO on the few held-out samples allows us to gain useful insights into the realistic performance of this method in a practical setting.
A generalization gap in the range between 3\% and 15\% is to be expected on the domain of CIFAR when replacing the test set \cite{recht2019imagenet}.
In general, we should not consider the best accuracy in \cref{subfig:gridsearch-cosine} as the optimal performance since we would be optimizing hyper-parameters on the test set.

To avoid the possibility of having methods ``luckier'' than others concerning HPO, for each training split, we run an independent hyper-parameter search.
Due to the comparatively small size of the validation sets in our benchmark, unstable HPO is not completely unlikely.
However, the validation sets cannot be much larger, since we operate in data-deficient settings and a sufficient number of samples needs to be available for the actual training.
We hence argue that our solution to this issue, \ie, averaging the results stemming from three groups of found hyper-parameters, improves the robustness of the results by considering the randomness of the search process.
Clearly, the more precision is requested for the estimate, the more the cost for evaluating a method on our benchmark.
We believe that increasing the number of HPO runs would exclude research groups without access to large clusters.
We have shown in \cref{tab:baseline-comparison} that our results for four methods originally evaluated in similar settings are either on par with the performance reported in the original publication or even outperform it.
This indicates that our searches found hyper-parameters as effective as the ones in the current literature.

\section{Conclusions}
\label{sec:conclusions}

We presented the first comprehensive overview and dedicated benchmark for \emph{deep learning from small datasets} in the context of image classification.

First, we carefully searched the literature for specialized methods applied to small-data tasks.
We categorized this collection into a constructive taxonomy and provided an overview to consolidate this field, which is currently very fragmented.

Second, analyzing our literature review, we found that a common evaluation benchmark with fixed datasets, architectures, and training pipelines was lacking in the research domain.
In addition, we found experimental evidence of weak baseline evaluations due to a lack of careful tuning.
To address the urge for a fair comparison, we developed a benchmark consisting of five datasets from different domains and data types.
Re-evaluating ten selected state-of-the-art methods led us to the surprising and sobering finding that standard cross-entropy loss is only surpassed by Harmonic Networks, and that performance growth is currently lacking in the literature.

In light of these results, we conclude that the importance of hyper-parameter optimization is immensely underestimated and should be considered in future studies to avoid misleading comparisons of new approaches with weak and underperforming baselines.
We also provide the largest collection of approaches to enable extensive comparisons with the state of the art.
We hope that our benchmark and training procedure will provide a fruitful basis for future developments and accelerate the progress in the field of image classification with small datasets.




\newpage
\printbibliography

\begin{IEEEbiography}[{\includegraphics[width=1in,height=1.25in,clip,keepaspectratio]{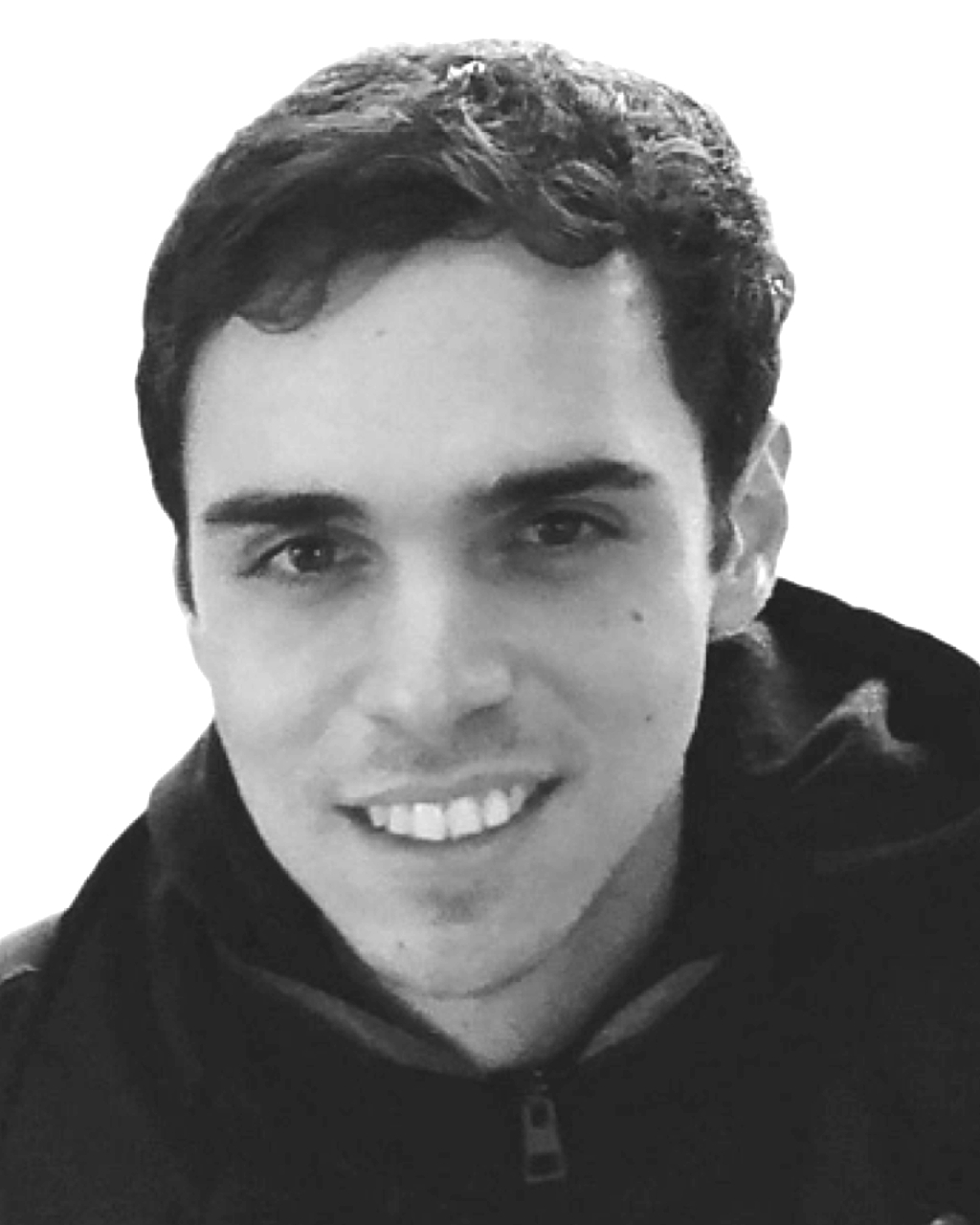}}]{Lorenzo Brigato} is a PhD candidate of the PhD Program in Engineering in Computer Science, and a member of the RoCoCo laboratory at the Department of Computer, Control and Management Engineering of the Sapienza University of Rome, Italy.
He earned an M.sc. in artificial intelligence and robotics with honours in 2018 from the same university.
His main research interests are in the field of data-efficient deep learning, anomaly detection, and robot learning.
\end{IEEEbiography}

\begin{IEEEbiography}[{\includegraphics[width=1in,height=1.25in,clip,keepaspectratio]{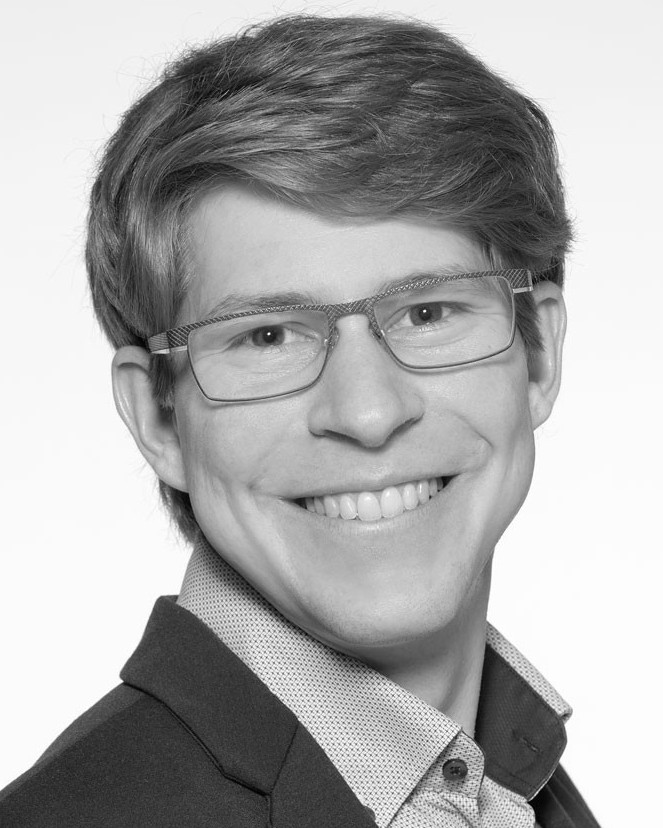}}]{Bj{\"o}rn Barz}
earned the M.Sc. in computer science with honours in 2016 from Friedrich Schiller University Jena, Germany, and received his PhD in 2021 with summa cum lauda for his work on semantic and interactive content-based image retrieval, which was done under the supervision of Joachim Denzler.
He was leading the knowledge integration team as a senior researcher at the Computer Vision Group Jena from 2020 to 2021 and is now machine learning scientist at Carl Zeiss AG.
His research interests are in the field of data-efficient deep learning, content-based image retrieval, and multi-modal learning.
\end{IEEEbiography}

\begin{IEEEbiography}[{\includegraphics[width=1in,height=1.25in,clip,keepaspectratio]{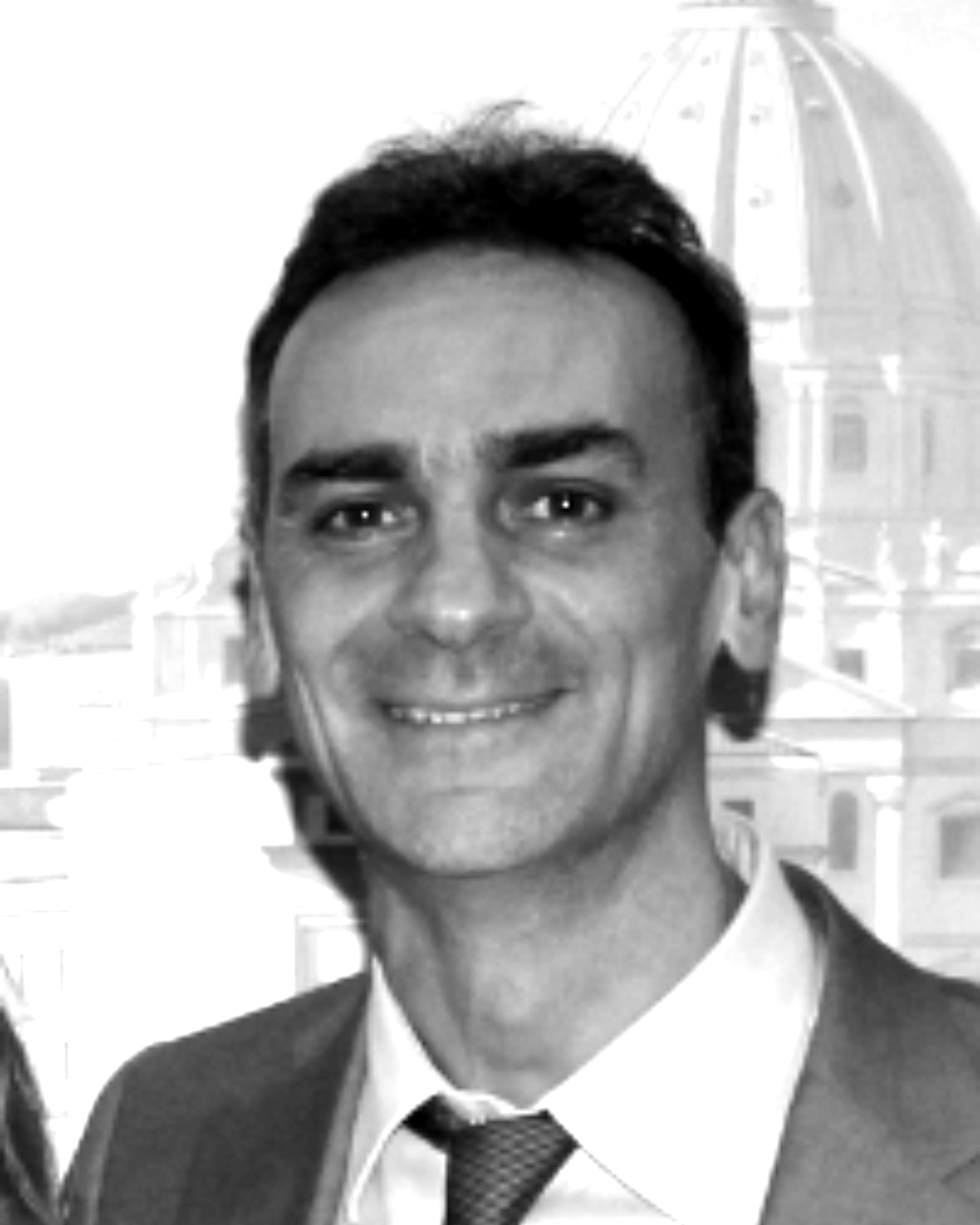}}]{Luca Iocchi} (www.diag.uniroma1.it/iocchi) is Full Professor at Sapienza University of Rome, Italy, mainly teaching in the Master in Artificial Intelligence and Robotics and currently Coordinator of the PhD Program in Engineering in Computer Science. 
His main research interests include cognitive robotics, task planning, multi-robot coordination, robot perception, robot learning, human-robot interaction, and social robotics.
He is the author of over 175 referred papers (h-index 43 [Google Scholar]) in journals and conferences in artificial intelligence and
robotics.
He has been principal investigator of several international, EU, national and industrial projects in artificial intelligence and robotics.
He is currently Vice-President of RoboCup Federation and organized several international scientific robot competitions, as well as student competitions focusing on service robots and human-robot interaction.
\end{IEEEbiography}

\begin{IEEEbiography}[{\includegraphics[width=1in,height=1.25in,clip,keepaspectratio]{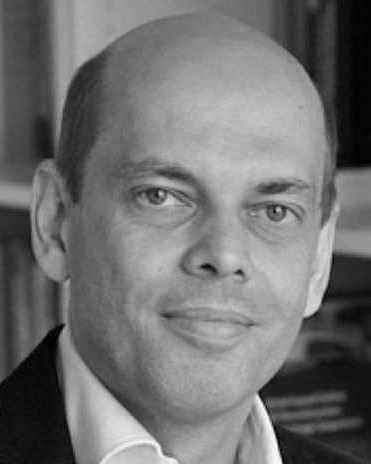}}]{Joachim Denzler}
(Member, IEEE) received the ``Diplom-Informatiker'', ``Dr.-Ing.'', and ``Habilitation'' degrees from the University of Erlangen, Germany, in 1992, 1997, and 2003, respectively.
He currently holds a position as full professor of computer science and the head of the 
Computer Vision Group, Friedrich Schiller University Jena, Germany.
He is also the director of the Data Science Institute of the German Aerospace Center (DLR) in Jena.
He is the author and coauthor of over 300 journal articles and conference papers as well as technical articles.
His research interests include automatic analysis, fusion, and understanding of sensor data, especially the development of methods for visual recognition tasks and dynamic scene analysis.
He contributed in the area of active vision, 3D reconstruction, and object recognition and tracking.
He is also a member of IEEE Computer Society, DAGM, and GI.
\end{IEEEbiography}

\EOD

\end{document}